\documentclass[11pt]{article}
\usepackage{amsmath, amssymb}
\usepackage{mathtools}
\usepackage{algpseudocode}
\usepackage{subcaption}
\usepackage{hyperref}
\usepackage{tikz}
\usepackage[most]{tcolorbox}
\usetikzlibrary{positioning, arrows.meta, calc}
\usepackage{float}
\usepackage[a4paper, total={5.86in, 9in}]{geometry}
\usetikzlibrary{positioning, shapes, calc}
\usepackage[linesnumbered,ruled,vlined]{algorithm2e}
\SetKw{KwDownTo}{down to}
\SetKwInOut{KwIn}{Input}
\SetKwInOut{KwOut}{Output}
\SetKw{KwBy}{by}
\usetikzlibrary{shapes,arrows, intersections}
\usepackage{caption}
\usepackage{natbib}

\title{High-Order Deep Meta-Learning with Category-Theoretic Interpretation}
\author{ David H. Mguni  \\
      Queen Mary University of London,\\ Peter Landin Building,\\ Mile End Road, \\London E1 4FZ}
\date{}
\begin{document}
\maketitle

\begin{abstract}
\noindent We introduce a new hierarchical deep learning framework for recursive higher-order meta-learning that enables neural networks (NNs) to construct, solve, and generalise across hierarchies of tasks. Central to this approach is a generative mechanism that creates \emph{virtual tasks}---synthetic problem instances designed to enable the meta-learner to learn \emph{soft constraints} and unknown generalisable rules across related tasks. Crucially, this enables the framework to generate its own informative, task-grounded datasets thereby freeing machine learning (ML) models from the limitations of relying entirely on human-generated data for training. By actively exploring the virtual point landscape and seeking out tasks lower-level learners find difficult, the meta-learner iteratively refines constraint regions. This enhances inductive biases, regularises the adaptation process, and can produce novel, unanticipated tasks and soft constraints required for generalisation. Each meta-level of the hierarchy corresponds to a progressively abstracted generalisation of problems solved at lower levels, enabling a structured and interpretable learning progression. This facilitates automatic curriculum construction, where higher meta-levels encode generalisations over families of tasks without requiring explicit specification. By interpreting meta-learners as category-theoretic \textit{functors} that generate and condition a hierarchy of subordinate learners, we establish a compositional structure that supports abstraction and knowledge transfer across progressively generalised tasks. The category-theoretic perspective unifies existing meta-learning models and reveals how learning processes can be transformed and compared through functorial relationships, while offering practical design principles for structuring meta-learning. 
We speculate this architecture may underpin the next generation of NNs capable of autonomously generating novel, instructive tasks and their solutions, thereby advancing ML towards general artificial intelligence.
\end{abstract}

\section{Introduction}

Meta-learning, often described as \emph{learning to learn}, has emerged as a powerful paradigm for developing adaptive systems that can rapidly acquire new skills from limited data \cite{vilalta2002perspective,hospedales2021meta,huisman2021survey}. Traditional approaches to meta-learning typically focus on a single level of adaptation, where a meta-learner optimises a base learner's initialisation or update rules to perform well across a distribution of tasks. However, many real-world problems require hierarchical learning strategies that extend beyond one level of adaptation, involving multiple layers of abstraction and generalisation \cite{bensusan2000higher}. Such \emph{higher-order} meta-learning poses significant challenges in terms of model design, optimisation, and theoretical understanding.

In this paper, we propose a novel recursive framework for higher-order meta-learning inspired and formalised through the lens of category theory~\cite{MacLane1998,leinster2014basic,awodey2010category}. Category theory provides a rich mathematical language for describing compositional structures and relationships between objects and morphisms, making it an ideal tool for modelling the complex interdependencies inherent in multi-level learning systems. By representing each meta-learner as a functor that maps learners at one level to learners at a more abstract level, we construct a principled hierarchical pipeline that captures the essence of abstraction and transfer in learning. This categorical viewpoint unifies diverse meta-learning techniques, revealing their common algebraic structure and guiding the design of more flexible and interpretable learning architectures.

A key feature of our approach is the incorporation of \emph{soft constraints} within the learning objectives, implemented through virtual data  similarly to physics-informed neural networks (PINNs)~\cite{karniadakis2021physics,Raissi2019}. These constraints regularise the adaptation of lower-level learners, ensuring consistency and smoothness across task distributions and facilitating knowledge transfer between meta-levels. This mechanism enables the meta-learners not only to optimise performance on observed tasks but to also generalise beyond the training distribution by leveraging structured knowledge embedded in the constraints.

Moreover, our framework naturally aligns with the principles of curriculum learning~\cite{graves2017automated,bengio2009curriculum}, where tasks are presented in an order that gradually increases in difficulty or abstraction. Each meta-level can be interpreted as a generalisation of the problems addressed at the level below, thereby organising the learning process as a progression from specific to more abstract problems. This perspective provides clarity on the role of curriculum design in meta-learning and offers practical guidance for constructing multi-layered learning pipelines that can tackle increasingly complex domains.

Potential applications of this hierarchical meta-learning framework span a broad spectrum, including continual learning~\cite{wang2024comprehensive}, automated scientific discovery~\cite{lu2024ai}, robotics~\cite{liu2022robot}, and control systems~\cite{bishop2011modern}, where adaptability and generalisation across tasks are crucial. By enabling neural networks to recursively learn abstract strategies and generate novel virtual tasks for self-supervision, our method paves the way towards systems capable of autonomous problem formulation and solution discovery. We speculate that such category-theoretic meta-learning architectures represent a foundational step towards truly general artificial intelligence, where machines not only solve given problems but also invent new ones to advance their knowledge and capabilities.

Training artificial intelligence (AI) models such as foundation models and large language models purely on human-generated data imposes inherent limitations on their capacity for generalisation, creativity, and autonomous abstraction. These models are bound by the statistical regularities and biases of their training corpora and lack mechanisms for generating new types of problems or concepts that fall outside the human-provided distribution. The recursive meta-learning framework addresses this by equipping higher-level learners with the ability to construct and explore synthetic tasks, constraints, or objectives that challenge and reshape the learning of subordinate levels. This process goes beyond imitation and enables models to autonomously generate instructive experiences, thereby escaping the confines of static datasets. In doing so, the framework lays groundwork for self-refining, curriculum-driven AI systems capable of discovering novel abstractions and adapting to out-of-distribution challenges, with profound implications for general intelligence.

\subsection*{Summary of Contributions}

This work introduces a unified, multi-level framework for recursive meta-learning, formalised using tools from category theory and realised concretely through differentiable neural architectures. It offers both a theoretical model of abstraction and constraint propagation across learning layers, and a practical template for implementing our framework across a range of neural network architectures.

\textit{1. Recursive Learning \&  Meta-Learning Generalisation.} 
Our core contribution is the development of a hierarchical learning architecture in which each level of learner acts over the structure of the level below, supplying it with soft constraints or virtual tasks to guide adaptation. These learners are recursive in nature; meta-learners constrain task learners, meta-meta-learners constrain meta-learners, and so on. The framework supports the compositional construction of learning systems where abstraction, control, and generalisation are explicitly modelled. Moreover, it generalises meta-learning models by offering a structured, semantically grounded framework for the learning of learning. 

 \textit{3. Adaptive exploration of the virtual point landscape.}
In our framework,  rather than relying on predefined soft-constraints or externally specified structures, the meta-learner explores the virtual point manifold by probing its lower-learner's failure modes by generating candidate tasks the lower-level learner finds difficult. Using this the meta-learner iteratively generates soft constraints and refines the subregion boundaries within which the soft constraints are expected to hold. 

\textit{4. Concrete Neural Network Realisation.}
In Section \ref{sec:our_framework}, we present an instantiation of the full framework, in which base learners, constraint generators, and meta-generators are parameterised by differentiable models trained using gradient descent. Each level propagates feedback through both supervised and virtual losses, enabling end-to-end optimisation of deeply nested learning objectives. This design ensures compatibility with modern autodiff systems while supporting flexible modularity in learner instantiation.

\textit{5. Category-Theoretic Formalisation.} 
In Section \ref{sec:category_th}, we formalise the architecture using category theory, where learning systems at each level are modelled as objects in categories, and their transformations as functors. This abstraction provides precise semantics for the relationships between tasks, learners, and constraints, and supports the recursive application of functorial structure to represent meta-level mappings. By drawing on category-theoretic ideas, we characterise how learners are fully determined by their relationships to other learners via transformations and constraints, suggesting a deep equivalence between structure and behaviour that informs both theory and design. This perspective provides a formal rationale for transformations and virtual points as the basis of abstract inference, elevating meta-learning from an empirical tool to a theoretically principled learning paradigm.

\textit{6. Applications Across Problem Domains.}
While our framework is domain-agnostic, in Section \ref{sec:abstraction} we illustrate its versatility through applications to multi-agent learning in game
-theoretic settings and reinforcement learning, demonstrating how nested learners can infer increasingly abstract solution structures, such as equilibria in structured games, without explicit semantic labelling. 

\section{Our Framework}
\label{sec:our_framework}
In this section, we introduce the key ideas underlying our framework and describe the main details of the architecture. We develop the framework pedagogically, first initiating the development of our framework by constructing the hierarchical design which produces a  novel framework for meta-learning. We  then describe the workings of each layer and architectural design with concrete instantiations. This leads to our first version of our meta-learning framework whose workings are encoded in Algorithm \ref{alg:recursive-meta}.  Thereafter, we introduce a set of techniques for performing exploration and generating virtual tasks for probing the lower level learners' capacity to solve specific tasks and for enabling the meta-learner to discover precisely structured soft-constraints. This leads to the full stack, multi-layer version of our framework with adversarially generative exploration whose workings are encoded in Algorithm \ref{alg:recursive-meta_adv_gen}.  In Section \ref{sec:category_th}, we analyse the framework through the lens of category theory and in so doing, provide insights about the framework's design principles and the structural relationships between its elements.      

Higher-order meta-learning aims to optimise models and learners and, the learning processes themselves recursively, and in a compositional manner~\cite{conklin2021meta}. We propose a principled framework based on \textit{category theory}~\cite{MacLane1998,awodey2010category}, which offers abstractions for composition, modularity, and transfer of learning procedures. 
Therefore, the framework we propose offers a principled approach to constructing recursive, higher-order meta-learning systems, using the compositional structures of category theory to model learners and learning processes. At its core, this framework treats learning algorithms not merely as black-box optimisers, but as composable mappings between spaces of tasks and models. In this way, it allows the modular design of learning systems where each layer of the stack learns to optimise the learning process of the layer below. 

The structure of our framework is organised as a hierarchy. In what follows, we make use of the category-theoretic notion of a \textit{functor}— a map between two systems of objects and relationships, in a way that preserves how relationships compose. At the lowest level of the hierarchy (level 0), individual models are trained on specific tasks drawn from a category of tasks \( \mathcal{T} \). These models are parameterised and optimised using a conventional loss function defined on the data for each task. At level \( k-1 \), we define a learner, represented as a functor \( L_{\phi_{k-1}} : \mathcal{T} \to \mathcal{M} \), mapping tasks to models in a structured and parameterised manner. The parameters \( \phi_{k-1} \) define how this learner adapts to each new task. Moving upward, a meta-learner \( F_k \) is defined as a functor that transforms learners, effectively learning how to improve the learning process itself. At each level \( k \), a meta-learner is promoted to become the learner at level \( k+1 \), resulting in a recursive stack where each layer learns to optimise the behaviour of the layer below.

One of the central innovations of this framework is the incorporation of \textit{soft
constraints} inspired by PINNs~\cite{karniadakis2021physics,cuomo2022scientific}. In particular, we propose that \textit{PINN-inspired soft constraints} be used to impose structure at each level of the hierarchy, by generating virtual tasks and corresponding loss terms, in direct analogy with how physical residuals guide learning in PINNs. In PINNs, physical laws are imposed as soft penalties in the loss function, evaluated on collocation points in the domain of the problem. We adopt a similar approach here, but instead of physical constraints, we impose \emph{meta-constraints} on the behaviour of learners and meta-learners. These constraints are enforced by generating virtual tasks or data points \( \tilde{T} \), drawn from a distribution defined by the desired meta-properties. The meta-learner is trained not only to minimise the empirical loss on real tasks, but also to minimise these additional penalty terms evaluated on the virtual tasks.

\subsection{Architecture}
\label{subsec:architecture}
In practice, this framework can be implemented using standard deep learning tools such as PyTorch~\cite{imambi2021pytorch} or JAX~\cite{bradbury2021jax}. Each learner and meta-learner can be parameterised as a neural network, with the meta-learner network outputting the parameters of the learner network at the layer below. The training procedure is structured as a nested optimisation loop. For each batch of real tasks, the learner is instantiated by the current meta-learner, trained on the task data, and evaluated on a hold-out set to compute the empirical loss. In parallel, virtual tasks \( \tilde{T} \) are sampled, either via known transformations (such as group actions, domain shifts, or controlled augmentations) or through a learned generative model conditioned on the current state of the learner. For these virtual tasks, synthetic data are constructed, and the learner's behaviour on these data is evaluated to compute the soft constraint loss.

To give concrete examples, one may enforce equivariance constraints\footnote{Equivariance is the property of a mapping such that transformation applied to the input produces an equivalent
transformation in the output~\cite{vander2022}.}~\cite{vander2022} by sampling group actions \( g \) and ensuring that the learner commutes with these actions. In domain adaptation settings, virtual tasks may represent shifted domains, and the meta-learner can be trained to produce learners whose performance remains stable across these shifts. In reinforcement learning~\cite{sutton1999reinforcement}, virtual environments with varied dynamics may be generated, enabling the meta-learner to produce learners that generalise across families of Markov decision processes (MDPs). 

\begin{figure}[ht]
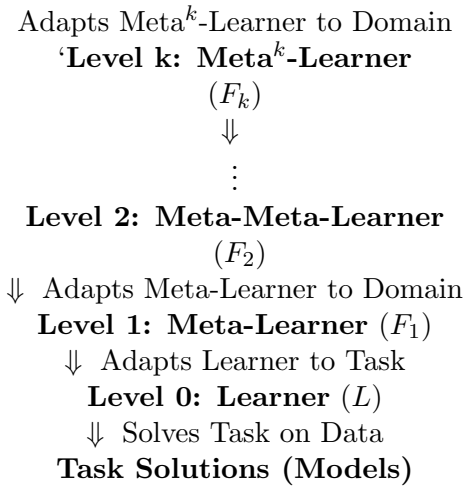

\[
\begin{array}{c}
\\
\text{Adapts Meta$^{k}$-Learner to Domain}\\`
\textbf{Level k: Meta$^k$-Learner}\\
(F_k)\\
\Downarrow \ \\ 
   \vdots 
\\
\textbf{Level 2: Meta-Meta-Learner}\\ \ (F_2) \\
\Downarrow \ \text{Adapts Meta-Learner to Domain} \\
\textbf{Level 1: Meta-Learner} \ (F_1) \\
\Downarrow \ \text{Adapts Learner to Task} \\
\textbf{Level 0: Learner} \ (L) \\
\Downarrow \ \text{Solves Task on Data} \\
\textbf{Task Solutions (Models)} \\
\end{array}
\]
\caption{\small Architecture of the hierarchical meta-learning framework. Each level corresponds to a progressively higher abstraction over the learning process. The base learner adapts to individual task instances. The meta-learner generalises across tasks within a domain. The meta-meta-learner abstracts across families of domains, learning how to adapt the meta-learning process itself. This constitutes an \emph{abstraction curriculum}, in which the system progressively acquires and transfers knowledge across increasingly abstract levels of representation. The full stack allows for category-theoretic analysis, treating each meta-level as a functor over the structures of the level below.}
\label{fig:stack_diagram}
\end{figure}

 The stack diagram in Figure \ref{fig:stack_diagram} illustrates the recursive structure of the multi-level meta-learning framework. Each level adapts the learning process at the level below, using functorial mappings. The injection of soft constraints via virtual tasks ensures that the system can enforce desirable properties and biases at each level of abstraction. Virtual data points allow the meta-learner and meta-meta-learner to generalise these constraints to new domains and tasks in a principled manner.
\begin{figure}[t]
\centering
\begin{tabular}{|c|c|p{8cm}|}
\hline
\textbf{Level} & \textbf{Component} & \textbf{Description} \\
\hline
$k+1$ & Meta-meta-learner $F_{k+1}$ & Learns to improve the meta-learner $F_k$ by optimising how the meta-learning process itself is structured. \\
\hline
$k$ & Meta-learner $F_k$ & Learns to improve the learner $L_{\phi_{k-1}}$ by mapping learners to improved learners or parameter updates. \\
\hline
$k-1$ & Learner $L_{\phi_{k-1}}$ & Maps tasks $T \in \mathcal{T}$ to models \( f_{\theta} \in \mathcal{M} \), adapting parameters $\phi_{k-1}$ to each task. \\
\hline
$0$ & Model $f_{\theta}$ & The base model trained directly on data for a specific task, optimised with task-specific loss. \\
\hline
\end{tabular}
\caption{\small Recursive higher-order meta-learning pipeline with PINN-inspired soft constraints and virtual data points. Each meta-learner is promoted to a learner at the next level. Soft constraints act via virtual tasks at each level.}
\label{fig:meta_pipeline_table}
\end{figure}
Soft constraints (virtual tasks) are incorporated between each layer, enforcing meta-properties by applying penalty terms such as $
\mathcal{L}_\text{virtual}(F_k(F_{k-1}), \tilde{T})$ and $ \mathcal{L}_\text{virtual}(L_{\phi_{k-1}}, \tilde{T})$, 
where \(\tilde{T}\) denotes virtual or synthetic tasks designed to induce invariances or robustness.

The recursive meta-learning framework may be instantiated in practice through a hierarchical composition of neural networks, each corresponding to a distinct level of abstraction in the learning process. At each level, the primary unit is a differentiable learner parameterised by a neural network, and information flows upward and downward through the stack via gradients and virtual tasks, enabling end-to-end optimisation with automatic differentiation.

At Level 0, the \emph{base learner} is a conventional task-specific neural model, such as a multi-layer perceptron (MLP)~\cite{riedmiller201}, convolutional neural network (CNN)~\cite{o2015introduction}, or transformer~\cite{vaswani2017attention}, depending on the input domain. This learner takes raw data $x$ and outputs a task-relevant prediction $y = f_{\theta_0}(x)$, where $\theta_0$ are trainable parameters. The model is trained via standard supervised or reinforcement learning objectives. Crucially, the training process itself is parameterised and modifiable by the upper levels.

At Level 1, the meta-learner governs how the base learner adapts to new tasks. One realisation is a \emph{hypernetwork} $g_{\theta_1}$, which takes a task embedding $z$ (e.g., derived from a task descriptor or a few data samples) and outputs the weights or initialisation $\theta_0 = g_{\theta_1}(z)$ for the base learner. Alternatively, $g_{\theta_1}$ may generate learning rates, loss weightings, or optimiser updates. This level captures task-level inductive biases and supports few-shot adaptation through parameter modulation. Training is done by computing a meta-loss $\mathcal{L}_1(\theta_1)$ that evaluates base learner performance after inner-loop updates.

At Level 2, the \emph{meta-meta-learner} operates on the space of task distributions or learning dynamics. It can be implemented as a generative model $h_{\theta_2}$ such as a variational autoencoder~\cite{pin2021}, diffusion model~\cite{atwood2016diffusion}, or transformer-based sampler~~\cite{vaswani2017attention} that generates synthetic tasks or entire learning episodes. The output of $h_{\theta_2}$ may take the form of datasets, task descriptors, or latent trajectories, which are then consumed by the meta-learner. The meta-meta-learner is trained by maximising downstream performance across many such generated tasks, defining a meta-meta loss $\mathcal{L}_2(\theta_2)$ based on aggregated evaluation of meta-learned models.

Backpropagation is performed through the entire hierarchy using automatic differentiation. This includes unrolling optimisation steps in the inner and outer loops (e.g., using truncated backpropagation through time), and differentiating through generated parameters, samples, or loss surfaces. In practical terms, this architecture may be implemented in frameworks such as PyTorch, JAX, or TensorFlow using nested gradient tapes or higher-order optimisers. 
This general architecture provides a modular and extensible foundation for recursive learning systems. By nesting learners, each trained to guide or shape the next, it realises an abstract structure in which inductive biases, learning curricula, and data generation strategies are all optimised jointly through experience, giving rise to dynamic, self-improving models.

\paragraph{Pipeline Architecture.}

At each level \( k \), we impose not only task-based losses but also \textit{soft constraints} through \textit{virtual tasks} \( \tilde{T} \) generated from the constraint. In particular, the loss functions can be specified by the following:
\\Level 0: Model \( f_\theta \)
\[
\mathcal{L}_\text{task}(f_\theta, T)
\]
\begin{center}
    \[\vdots\]
\end{center}
\noindent{Level \( k-1 \): Learner \( L_{\phi_{k-1}} \)}
\[
\mathcal{L}_\text{learner}(\phi_{k-1}) = \mathbb{E}_{T} \left[ \mathcal{L}_\text{task}(f_{\theta^*}, T) \right] + \lambda_{k-1} \cdot \mathcal{L}_\text{virtual}(L_{\phi_{k-1}}, \tilde{T})
\].

\noindent{Level \( k \): Meta-learner \( F_k \)}
\[
\mathcal{L}_k(\xi_k) = \mathbb{E}_{T} \left[ \mathcal{L}_{k-1}(F_k(F_{k-1}), T) \right] + \lambda_k \cdot \mathbb{E}_{\tilde{T}} \left[ \mathcal{L}_\text{virtual}(F_k(F_{k-1}), \tilde{T}) \right].
\]

The recursive promotion process involves, at each level, the meta-learner \( F_k \) becoming the learner for the next higher-order task.
 \label{fig:architecure_2}
\begin{figure}[t]
\centering
\begin{tikzpicture}[
    node distance=1.8cm and 3.5cm,
    every node/.style={draw, rounded corners, align=center, minimum width=3.5cm, minimum height=1cm, font=\scriptsize},
    arrow/.style={-{Latex[length=3mm]}, thick} 
]
\node (Lk1) {Level $k+1$ \\ Meta-meta-learner $F_{k+1}$};
\node[below=of Lk1] (Lk) {Level $k$ \\ Meta-learner $F_k$
\\Learns $\mathcal{C}_{\text{soft}}(\phi_k)$ + $L_k$};
\node[below=of Lk] (Lk_1) {Level $k-1$ \\ Learner $L_{\phi_{k-1}}$
\\Trained on $(T_{k-1} \cup \tilde{T}_{k-1})$ };
\node[below=of Lk_1] (Model) {Level 0 \\ Model $f_{\theta}$ \\ Trained on $T$};

\draw[arrow] (Lk1) -- (Lk) node[midway, right] {Learner promotion};
\draw[arrow] (Lk) -- (Lk_1) node[midway, right] {Learner promotion};
\draw[arrow] (Lk_1) -- (Model) node[midway, right] {Learner promotion};
\draw[arrow, dashed, bend left=25] (Lk) to node[midway, left, xshift=-3mm] {\small Soft constraints \\ Virtual tasks $\tilde{T}_{k-1}$} (Lk_1);
\draw[arrow, dashed, bend left=25] (Lk_1) to node[midway, left, xshift=-3mm] {\small Soft constraints \\ Virtual tasks $\tilde{T}_0$} (Model);
\end{tikzpicture}
\caption{\small Recursive higher-order meta-learning pipeline with PINN-inspired soft constraints. Each meta-learner is promoted to learner at next level. The meta-learners generate virtual points to regularise lower levels. The soft constraints act via virtual tasks at each level. }
\label{fig:meta_pipeline}
\end{figure}
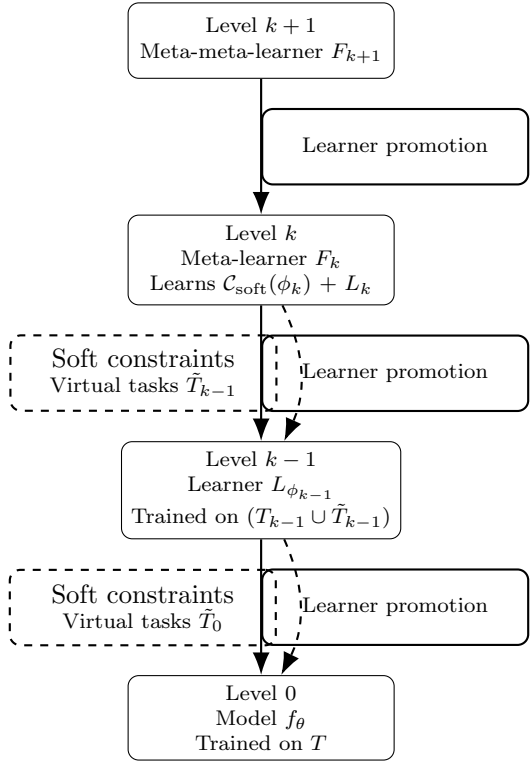
\paragraph{Soft Constraints and Virtual Data.}
Soft constraints are expressed as penalty terms evaluated on \textit{virtual tasks} \( \tilde{T} \). This is analogous to collocation points in PINNs~\cite{karniadakis2021physics}. 

A key mechanism in our recursive higher-order meta-learning pipeline is the generation of \emph{virtual points} or \emph{virtual tasks} through soft constraints learned at each meta-level. Virtual points are synthetic data points or tasks generated by higher-level meta-learners to regularise and guide the training of lower-level learners via soft constraints. This process is depicted in Figure \ref{fig:meta_pipeline} which shows the meta-learning pipeline and the injection of soft constraints at lower levels. The virtual datasets can be interpreted as synthetic datasets that enforce inductive biases and enable the curriculum learning interpretation of the hierarchy. The flexibility to generate these points either directly or adversarially expands the applicability of this framework to a wide range of domains. The meta-learner learns to generate virtual tasks \( \tilde{T} \sim \mathcal{C}_\text{soft} \).
Just as in PINNs, where physics constraints act on synthetic points, here meta-constraints act on virtual tasks.

In our framework, at meta-level \( k \), the meta-learner learns both a meta-learning model \( L_k \) and a soft constraint model \( \mathcal{C}_{\text{soft}}(\phi_k) \). At meta-level \( k \), the meta-learner \( L_k \) learns a soft constraint model \( \mathcal{C}_{\text{soft}}(\phi_k) \) capturing inductive biases or abstract generalisations about desirable behaviours of lower-level learners \( L_{k-1} \). The soft constraint model allows the meta-learner to generate virtual datasets $
\tilde{T}_{k-1} = \{ (x_i, y_i) \}_{i=1}^N \sim \mathcal{C}_{\text{soft}}(\phi_k)$, which form a synthetic dataset used to supplement training data for learners \( L_{k-1} \). In this sense, virtual points at level \( k \) become new datasets for lower-level tasks \( T_{k-1} \). The meta-learner is then updated based on a composite meta-loss:
\[
\mathcal{L}_k = \mathbb{E}_{T_k} \left[ \mathcal{L}_{\text{task}}(f_{\theta_{k-1}^*}, T_{k-1}) + \lambda \mathcal{L}_{\text{virtual}}(f_{\theta_{k-1}^*}, \tilde{T}_{k-1}) \right],
\]
where \( \lambda \in\mathbb{R}_{\geq 0}\) balances the influence of real and virtual data. Two strategies are available for generating virtual points.
\newline\textbf{Direct Sampling.} If \( \mathcal{C}_{\text{soft}} \) is tractable (e.g. a known symmetry, differential equation, smoothness prior), we can sample valid virtual points directly from it.
\newline\textbf{Adversarial Generation.} If \( \mathcal{C}_{\text{soft}} \) is complex or implicitly defined, a GAN-like architecture~ \cite{goodfellow2020generative}  can be used with the following procedure:
    \begin{enumerate}
        \item A generator \( G_{\psi} \) proposes candidate virtual points.
        \item A discriminator \( D \) penalises points violating \( \mathcal{C}_{\text{soft}} \).
        \item The generator and meta-learner are trained jointly to ensure virtual points adhere to constraints.
    \end{enumerate}

This adversarial approach is particularly useful when constraints must be learned from data, such as in computer vision or natural language processing domains. This mechanism allows meta-learners to autonomously design curricula of virtual tasks that progressively abstract knowledge across levels of learning. Combined with the recursive architecture, it provides a pathway toward systems that can autonomously discover, abstract, and generalise solutions to new classes of problems.
\begin{tcolorbox}[%
    enhanced, 
    breakable,
    frame hidden,
    overlay broken = {
        \draw[line width=.4mm, black, rounded corners]
        (frame.north west) rectangle (frame.south east);},
    ]{}
\subsection*{Example 1: Curriculum Learning and Meta-Abstraction.} 
Consider a recursive pipeline learning mathematical transformations:
\newline\emph{Level 0:} Learns polynomial regression on datasets \( y = a x^2 + b x + c \).
\newline\emph{Level 1:} Learns abstract invariances and symmetries (e.g. scaling invariance).
\newline\emph{Level 2:} Learns that certain parameter regions (e.g. smooth polynomials) generalise better.

At level 2, the meta-learner encodes \( \mathcal{C}_{\text{soft}} \) as a prior over polynomial coefficients. It generates virtual polynomial tasks consistent with these priors, shaping level 1 learners to prioritise generalisable solutions.
\end{tcolorbox}

\subsection*{Algorithm}

\begin{algorithm}[t]
\caption{\small Recursive Higher-Order Meta-Learning with Soft Constraint Virtual Tasks}
\label{alg:recursive-meta}
\KwIn{Number of meta-levels $K$, meta-task distributions $p(\mathcal{T}_k)$, learning rates $\eta_k$}
\KwOut{Meta-learners $L_k$ with parameters $\xi_k$ and soft constraints $\mathcal{C}_{\text{soft}}(\phi_k)$}

\For{$k \leftarrow K$ \KwTo $1$ \KwBy $-1$}{
    \ForEach{batch of meta-tasks $T_k \sim p(\mathcal{T}_k)$}{
    
        Sample $F_{k-1}$ or $L_{\phi_{k-1}}$ using current $F_k$\;
        
        \ForEach{meta-task $T_k$}{
            \ForEach{task $T_{k-1} \in T_k$}{
                Instantiate learner $f_{\theta_i}$ via $L_{\phi_{k-1}}$\;
                Train $f_{\theta_i}$ on $T_{k-1}$ to obtain $f_{\theta_i^*}$\;
                Compute $\mathcal{L}_{\text{task}}(f_{\theta_i^*}, T_{k-1})$\;
            }
            
            \eIf{Direct sampling is feasible}{
                Sample $\tilde{T}_{k-1} \sim \mathcal{C}_{\text{soft}}(\phi_k)$\;
            }{
                Train generator $G_\psi$ against discriminator $D$ to enforce $\mathcal{C}_{\text{soft}}$\;
                Sample $\tilde{T}_{k-1}$ from $G_\psi$\;
            }
            
            Compute $\mathcal{L}_{\text{virtual}}(f_{\theta_i^*}, \tilde{T}_{k-1})$\;
        }
        
        Compute composite meta-loss $\mathcal{L}_k = \mathcal{L}_{\text{task}} + \lambda \mathcal{L}_{\text{virtual}}$\;
        Update meta-learner parameters:
        $\xi_k \leftarrow \xi_k - \eta_k \nabla_{\xi_k} \mathcal{L}_k$\;
    }
}
\end{algorithm}

    
    
        
        
    

Algorithm~\ref{alg:recursive-meta} formalises the general multi-layer hierarchical meta-learning procedure, where each meta-level $k$ acts as a meta-learner $F_k$ that generates or conditions the learner at the next lower level, $L_{\phi_{k-1}}$. The base-level models $f_{\theta_i}$ are adapted to individual tasks $T_i$ by minimising task-specific losses $\mathcal{L}_\text{task}$. Crucially, the meta-learner also enforces soft constraints through a set of \emph{virtual tasks} $\tilde{T}_j$ sampled from a learned constraint distribution $\mathcal{C}_\text{soft}$. This mechanism is analogous to virtual data points in physics-informed neural networks, where the constraints shape the solution space to improve generalisation and inductive bias transfer. By iterating this process recursively, higher-order meta-learners abstract general strategies from the adaptation dynamics of the lower levels, forming a compositional pipeline that can be analysed categorically. Each meta-level corresponds to a functorial mapping over the structure of the previous level, enabling principled reasoning about knowledge transfer and curriculum learning along the axis of abstraction.

\subsection{Exploration in the Virtual Task Space via Meta-Learner Discovery}\label{subsec:exploration_sec}

A fundamental challenge in recursive meta-learning architectures is the autonomous discovery of informative virtual tasks; synthetic or abstract data points that reveal weaknesses in subordinate learners. Particularly at intermediate levels of the hierarchy, where explicit inductive biases may be absent, a key responsibility of the meta-learner is to explore the space of virtual tasks so as to expose regions where the base learner struggles to generalise or adapt.

In what follows, we now conceptualise the meta-learner as an active explorer within a \emph{virtual point manifold} $\mathcal{V}$, tasked with generating candidate tasks or constraints $\tilde{x} \in \mathcal{V}$ that the base learner finds difficult. Rather than relying on predefined difficulty heuristics or externally specified structures (e.g. game payoff structures), the meta-learner discovers such instances by probing the failure modes of its base learner. We do this by maintaining a generative model $G_\phi: \mathcal{Z} \to \mathcal{V}$ over virtual tasks, parameterised by latent codes $z \in \mathcal{Z}$ and conditioned on historical training feedback. The generator is trained to maximise the expected difficulty or uncertainty as perceived by the lower-level learner.

\begin{figure}
    \centering
    \includegraphics[width=0.4\linewidth]{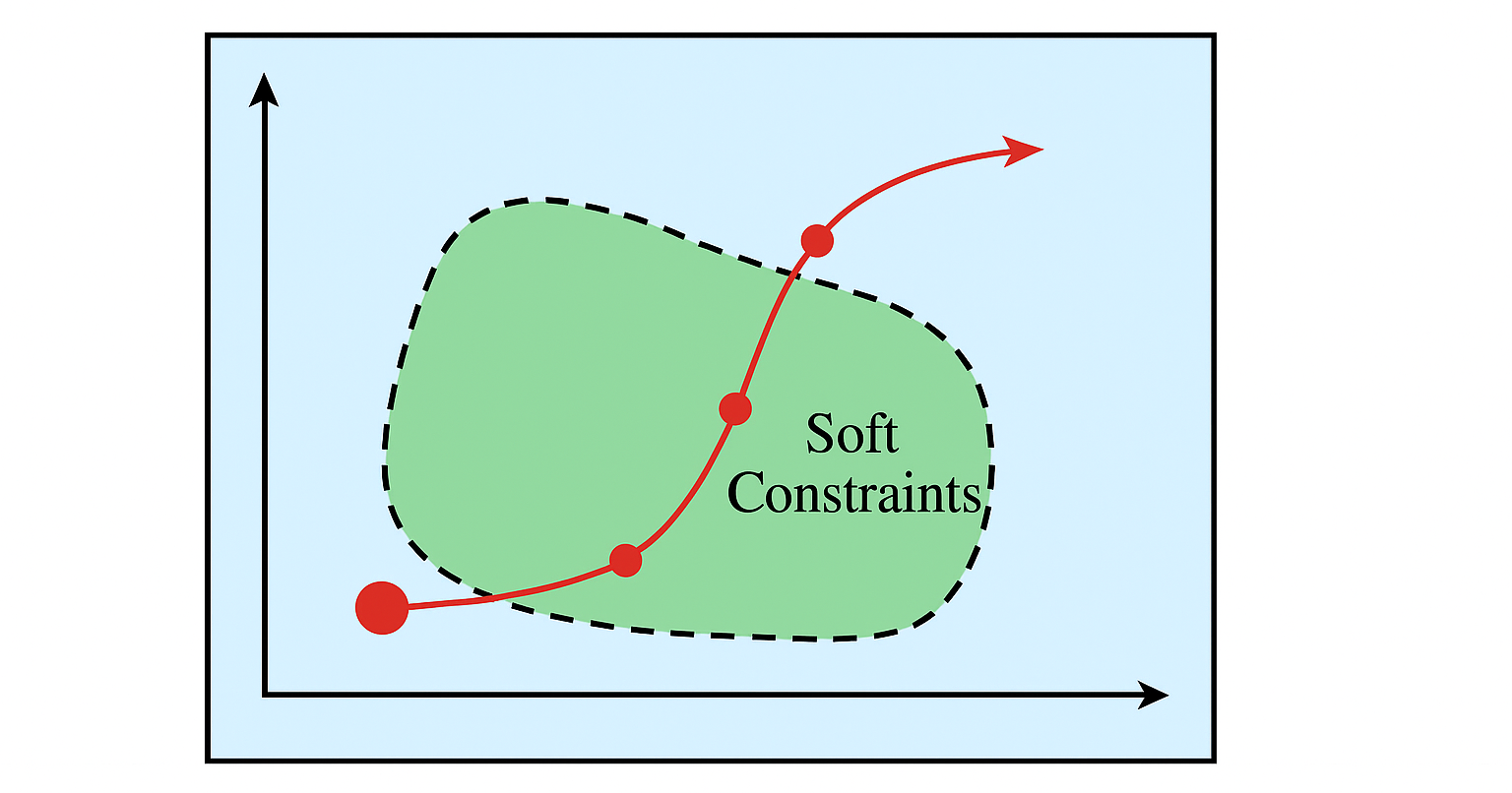}
    \caption{\small The virtual point landscape. The generative model can be used to explore the landscape and identify subregions where the soft constraints are valid and discover new soft constraints.}
    \label{fig:virtual_land_learner}
\end{figure}

Over successive meta-training iterations, the base learner improves, and the meta-learner must respond by generating increasingly abstract or compositionally novel virtual tasks. This process gives rise to an emergent curriculum; not one prescribed by a human designer, but shaped by the base learner's evolving competence. Crucially, this curriculum is not limited to known task types. This may give rise to novel forms of tasks or constraints that were not anticipated but that are necessary for generalisation.

As the recursive learner hierarchy progresses, the structure of virtual tasks becomes increasingly abstract. The virtual points constructed at higher levels may correspond not to specific datasets, but to equivalence classes of tasks (e.g., potential games, coordination regimes), providing a curriculum of generalisation at increasing levels of conceptual abstraction. These emergent categories reflect not just performance boundaries but deeper representational regularities, guiding learning through conceptual discovery.


We formalise the virtual task exploration mechanism employed by an intermediate or higher-level meta-learner within the recursive framework. This learner aims to identify virtual data points that expose limitations in the performance of the base-level learner, especially in the absence of explicit inductive biases.
%
Let the system be modelled via a hierarchy of transformations between tasks, learners, constraints, meta-problems, and higher-order abstractions. Each level of this hierarchy participates in the training and evaluation of learners over increasingly abstract domains.

Let $\theta$ denote the parameters of a base learner at level 0. Let $\phi$ denote the parameters of a virtual task generator $G_\phi: \mathcal{Z} \to \mathcal{V}$, mapping latent codes $z \in \mathcal{Z}$ to virtual tasks $\tilde{x} \in \mathcal{V}$,  $\mathcal{L}_{\text{task}}$ denote the base-level loss on a specific task. Lastly, denote by $\mathcal{L}_{\text{explore}}(\tilde{x}, \theta)$ a function quantifying the difficulty of virtual point $\tilde{x}$ with respect to the current learner parameters $\theta$. The meta-learner's goal is to find generator parameters $\phi$ such that the generated virtual points are maximally difficult for the base learner. The objective function for the generator is therefore given by:
\[
\max_{\phi} \ \mathbb{E}_{z \sim \mathcal{Z}} \left[ \mathcal{L}_{\text{explore}}(G_\phi(z), \theta) \right] \quad \text{subject to} \quad G_\phi(z) \in \mathcal{M}_{\text{valid}},
\]
where $\mathcal{M}_{\text{valid}}$ is a constraint manifold that restricts $G_\phi(z)$ to produce plausible data.

The resulting training loop is adversarial in nature; the base learner seeks to minimise the overall task loss over both real and virtual data, while the generator seeks to find virtual points that challenge this learner. 
The generator and discriminator play central roles in enabling active task-space exploration and the creation of synthetic learning signals that guide abstraction and generalisation across levels. The generator, $G_{\phi_k}$ at meta-level $k$ is responsible for producing virtual tasks or synthetic data points that are not present in the original training distribution. These virtual tasks are sampled to challenge the lower-level learner by exploiting regions of task-space where the learner either performs poorly or where generalisation behaviour is unclear or underdeveloped. The generator is trained to maximise an exploration reward that reflects the novelty, difficulty, and informativeness of its samples. This reward can incorporate contextual exploration scores that measure, for instance, the loss difference relative to nearby tasks or divergence from the manifold of previously visited problems.

The discriminator $D_{\psi_k}$ acts as an adversarial critic that estimates whether a generated task lies within the manifold of tasks that are feasible, meaningful, or aligned with previously learned constraints. It enforces plausibility, coherence, or constraint satisfaction in the generator's outputs. The adversarial training between $G_{\phi_k}$ and $D_{\psi_k}$ therefore defines a soft boundary for the valid task manifold, enabling the generator to probe near-boundary regions and thereby drive the discovery of soft constraints or new regularities. Together, the generator–discriminator pair enables higher-level meta-learners to sculpt the learning landscape of lower levels by autonomously generating instructive virtual tasks and adapting the constraints under which learning occurs. This dynamic also induces a natural curriculum as the generator adapts to the learner's capabilities. It contributes directly to the recursive abstraction of task structure and the construction of curriculum-like trajectories in task space, fundamental to the framework's ability to support general-purpose learning. At higher levels of the hierarchy, this interpretive mechanism becomes even more abstract. The meta-meta-learner, for instance, can generate virtual constraints or curricula that apply across entire families of meta-learning tasks. Here, virtual points operate in highly abstract spaces, and the constraints they help to define govern not just learners, but spaces of learners. The recursive abstraction naturally supports learning-to-learn dynamics across multiple orders, enabling the framework to autonomously discover inductive structure without direct supervision at each level. 
Integrating the adversarial virtual point generation mechanism within Algorithm \ref{alg:recursive-meta}, leads to our Recursive Meta-Learning with Virtual Exploration
Algorithm (Algorithm \ref{alg:recursive-meta_adv_gen}) which describes the workings of our finalised framework.

Having introduced an adversarial virtual point generation process whose goal is to find difficult tasks for the lower level learner, a natural question arises: 
\begin{center} 
\textit{\textbf{Should these virtual points always respect the soft constraints inferred by the meta-learner?} }
\end{center}

The answer, crucially, is negative. The virtual point generator serves a more flexible and expressive role, and this non-necessity is a fundamental strength of the framework.

At the base level, virtual points often act as test-time probes or robustness checks for the learner. These points may or may not conform to the currently inferred soft constraints. Their main purpose is to expose the inductive limitations of the base learner, especially in regions of the input space where the constraints offer weak guidance or where the training distribution offers insufficient coverage. The meta-learner, observing how the base learner responds to such points, can iteratively refine the constraint landscape.

At the meta-learning level, the learner is not just constrained by fixed soft rules but actively infers these constraints based on the loss signals induced by the virtual points. When virtual points violate existing constraints and result in poor generalisation, this discrepancy forms a powerful learning signal and can be used to ascertain vital information about the boundaries of the constraint landscape (depicted in Figure \ref{fig:virtual_land_learner}). Conversely, when virtual points align with constraint-satisfying behaviour, they provide positive evidence for the correctness of the inferred bias. Through repeated exposure and gradient-based adaptation, the meta-learner internalises these patterns, effectively distilling a soft inductive bias that guides the base learner across tasks. Therefore, even when virtual points fall outside the scope of current constraints, they contribute to the discovery of a constraint by illuminating its failure modes. 

\begin{algorithm}[H]\small
\caption{\small Recursive Meta-Learning with Constrained Virtual Exploration}
\label{alg:recursive-meta_adv_gen}
\KwIn{Meta-levels $k = 1$ to $K$, data distributions $\{\mathcal{D}_k\}$, training iterations $T$}
\KwOut{Trained meta-learners $\Phi_k$ and generators $G_{\phi_k}$ at each level}

\For{$k \leftarrow K$ \KwTo $1$}{
    Initialise meta-learner parameters $\Phi_k$ and generator parameters $\phi_k$, discriminator $D_{\psi_k}$\;
    
    \For{$t \leftarrow 1$ \KwTo $T$}{
        Sample batch of meta-tasks $\{\mathcal{T}_i\}_{i=1}^B$ from $\mathcal{D}_k$\;

        \ForEach{meta-task $\mathcal{T}_i$}{
            Sample or synthesise lower-level learner (or loss function) from $\Phi_k$\;

            \ForEach{task instance $\tau_{ij} \in \mathcal{T}_i$}{
                Instantiate learner $f_{\theta_{ij}}^{(k-1)}$ and train on $\tau_{ij}$\;
                Compute task loss $\mathcal{L}_{\text{task}}^{(ij)} = \mathcal{L}(f_{\theta_{ij}}, \tau_{ij})$\;
            }

            \ForEach{sampled latent code $z$}{
                Generate virtual task $\tilde{\tau}_{ij} = G_{\phi_k}(z)$\;
                Compute exploration score $\mathcal{S}_{\text{explore}}(\tilde{\tau}_{ij})$\;
                Compute manifold penalty $\mathcal{L}_{\text{manifold}}(\tilde{\tau}_{ij})$\;
                Compute discriminator loss: $\mathcal{L}_{\text{adv}} = -\log D_{\psi_k}(\tilde{\tau}_{ij})$\;
                Compute generator objective:
                \[
                \mathcal{L}_{\text{gen}}(\phi_k) = -\mathcal{S}_{\text{explore}}(\tilde{\tau}_{ij}) 
                + \beta \mathcal{L}_{\text{manifold}}(\tilde{\tau}_{ij}) 
                + \gamma \mathcal{L}_{\text{adv}}
                \]
            }

            Compute virtual loss $\mathcal{L}_{\text{virtual}}^{(ij)} = \mathcal{L}(f_{\theta_{ij}}, \tilde{\tau}_{ij})$\;
        }

        Compute total loss:
        \[
        \mathcal{L}_{\text{meta}} = \sum_{i,j} \left( \mathcal{L}_{\text{task}}^{(ij)} + \lambda \mathcal{L}_{\text{virtual}}^{(ij)} \right)
        \]

        Update meta-learner parameters: $\Phi_k \gets \Phi_k - \eta_\Phi \nabla_{\Phi_k} \mathcal{L}_{\text{meta}}$\;

        Update generator parameters: $\phi_k \gets \phi_k - \eta_\phi \nabla_{\phi_k} \mathcal{L}_{\text{gen}}$\;

        Update discriminator parameters: $\psi_k \gets \psi_k - \eta_\psi \nabla_{\psi_k} \log D_{\psi_k}(\tilde{\tau}_{ij})$\;
    }
}
\end{algorithm}

\vspace{.25cm}
\paragraph{Exploration Criterion.} 

A plausible criterion for exploration is the expected predictive entropy $\mathbb{H}[p_\theta(y | \tilde{x})]$, where $p_\theta$ denotes the base learner's output distribution on virtual input $\tilde{x}$. In this case the meta-learner seeks to generate virtual points that maximise this entropy, subject to regularity constraints ensuring semantic plausibility. Another set of possibilities are loss-gradient divergence metrics or validation losses can serve as exploration signals, 
predictive entropy, $\mathbb{H}[p_\theta(y \mid \tilde{x})]$, the loss, $ \mathcal{L}_{\text{task}}(f_\theta(\tilde{x}), y)$, and the gradient norm, $\left\| \nabla_\theta \mathcal{L}_{\text{task}}(f_\theta(\tilde{x}), y) \right\|$. These metrics are nonetheless limited in their scope to prioritise the generation of tasks that reveal important subregion boundaries in the soft constraint landscape.

To guide the generation of informative new tasks $\{\tau_i\}$ during meta-training in our framework, we introduce an exploration criterion that scores tasks according to their utility for improving generalisation.  A task is considered valuable for exploration if it exhibits either a \emph{sharp transition} in learner difficulty relative to a nearby reference task, or if it is \emph{surprisingly easy or difficult} relative to a distant one. This aims to expose discontinuities in performance or unexpectedly generalisable patterns.
 To this end, we propose a criterion that prioritises tasks which are either (i) significantly more or less difficult than nearby tasks, or (ii) exhibit stable performance across distant regions in task space. Let $\tau$ denote a candidate task, and $\tau_{\text{ref}}$ a reference task (e.g., one previously mastered or currently being learned). Define the \emph{difficulty gap} as $\Delta \mathcal{L}(\tau_\text{ref}, \tau) := \mathcal{L}(\theta, \tau) - \mathcal{L}(\theta, \tau_\text{ref})$, and let $d(\tau, \tau_{\text{ref}})$ be a task-space distance according to some metric (e.g., embedding or problem parameter space).

The contextual score is then given by:
\begin{equation}
\mathcal{S}_{\text{explore}}(\tau) =
\lambda \cdot \max\left(\alpha_1, \frac{\Delta \mathcal{L}(\tau_\text{ref}, \tau)}{d(\tau, \tau_{\text{ref}}) + \delta} - \epsilon_1 \right)
+ (1 - \lambda) \cdot \max\left(\alpha_2, \frac{d(\tau, \tau_{\text{ref}})}{1 + \lvert \Delta \mathcal{L}(\tau_\text{ref}, \tau) \rvert} - \epsilon_2 \right),
\end{equation}
where $\lambda \in [0,1]$ balances the two components, $\delta > 0$ is for numerical stability, and $\epsilon_1, \epsilon_2$ are are thresholds that suppress uninformative contributions. The terms $\alpha_1, \alpha_2 < 0$ act as penalties for tasks that do not meaningfully satisfy the criterion. 
The first term promotes tasks that cause large changes in learner performance over small distances (sharp local transitions), while the second term rewards tasks that are distant yet elicit similar loss (indicating stable generalisation). Given a current task $\tau$, we select reference tasks from the learner's visitation history $\mathcal{H}$ using extremal criteria i.e.,
$\tau_{\text{ref}} \in \arg\min_{\tau' \in \mathcal{H}} d(\tau, \tau')$. This structure ensures that a task contributes to the exploration signal only if it meaningfully satisfies at least one of the two objectives. Tasks that fail to exceed either threshold are automatically ignored. This allows for the discovery of sharp task-space boundaries as well as the identification of robust generalisation regimes, and enables principled sampling of tasks that are most informative for learner improvement. In practice, $\mathcal{L}(\theta, \mathcal{T})$ can be estimated via validation loss or episodic return, and $D(\mathcal{T}_\text{ref}, \mathcal{T}_i)$ can be computed in a learned task embedding space or using hyperparameter distance metrics.

To ensure that the generator produces structurally valid and learnable tasks, we introduce an additional constraint that encourages generated tasks to lie within a plausible task manifold. This can be enforced either through a soft constraint function $\mathcal{L}_{\text{manifold}}$ representing known inductive structure (e.g., smoothness, physical laws, symmetries), or via a discriminator $D_\psi$ that learns to differentiate real tasks from synthetic ones. After making the constraint explicit, the overall generator objective then becomes:
\begin{equation}
\mathcal{L}_{\text{gen}}(\phi) = 
- \mathbb{E}_{z}[\mathcal{S}_{\text{explore}}(G_\phi(z))] 
+ \beta \cdot \mathbb{E}_{z}[\mathcal{L}_{\text{manifold}}(G_\phi(z))]
- \gamma \cdot \mathbb{E}_{z}[\log D_\psi(G_\phi(z))],
\end{equation}
where $\beta$ and $\gamma$ control the influence of the soft constraint and adversarial discriminator, respectively. This regularisation encourages the generator to produce tasks that are both informative under the exploration criterion and consistent with prior structural knowledge or the observed distribution of real tasks. Incorporating such constraints improves the stability of the generator's training and enhances the semantic quality of the generated task distribution. The objective includes a number of components that rely on non-differentiable \texttt{max} operations that can lead to numerical instabilities. To remedy this, in Section \ref{sec:smooth_context}, we replace these components with functions that preserve the semantics of the objective while ensuring differentiability for stable gradient flow during optimisation. 

\textbf{Representing structural Relationships in the constraint manifold.} To encode abstract structural relationships within a learned constraint manifold, we turn to mathematical frameworks that capture transformation, compositionality, and invariance properties. Notably, algebraic and differential geometric structures such as \emph{Lie groups}, \emph{rings}, and their generalisations provide natural inductive biases and constraints for learning (see Section \ref{sec:maths_structure} for brief overviews on rings and Lie groups, for exhaustive discussions on the respective subjects we refer the reader to \cite{hall2013lie} and \cite{irving2004integers}). 

Architecturally, such structures can be integrated via 1. Symmetry-aware networks: Equivariant neural networks (e.g., group CNNs~\cite{pmlr-v48-cohenc16}, LieConv~\cite{finzi2020generalizing}) that maintain the structural properties throughout the computation. 2. Algebraic latent spaces: Encodings over ring elements, or Lie group embeddings, with learned group actions \cite{batatia2023general}. 3. Manifold-regularised objectives: Loss functions that penalise deviation from the manifold defined by the group or ring structure. Representing constraints via Lie groups or ring-like algebraic systems constrains the search space to meaningful and interpretable classes of hypotheses. This not only reduces overfitting and improves generalisation but also aligns well with the semantics of many domains such as physics, robotics, symbolic reasoning, and language. 
In our framework, such structures serve as prior inductive scaffolds that define the \emph{geometry} and \emph{algebra} of the constraint manifold, guiding the meta-learners in generating task distributions and constraints that preserve fundamental properties across levels of abstraction.

To enhance both the generalisability and efficiency of constraint learning within our recursive meta-learning framework, we propose the inclusion of multiple candidate structural representations such as Lie groups, rings, and general algebraic or geometric priors encoded within the generator or constraint modules. Rather than relying on a fixed structural form, the framework learns to selectively retain structures that prove consistently useful across tasks, while discarding those that fail to improve generalisation. This adaptive structural selection functions as a meta-level inductive bias discovery mechanism.

Concretely, we consider a collection of structural modules, each representing a candidate transformation law or compositional structure (e.g., equivariance under a Lie group action, distributive ring operations, or gauge transformations). These modules are conditionally activated by a learned selector function, defined as
\[
s_i(\tau) = \sigma(w_i^\top \phi(\tau)),
\]
where \( \phi(\tau) \) is an embedding of the current task, \( w_i \) are learnable weights for structure \( i \), and \( \sigma \) is a softmax or sigmoid gating function. This selector governs the contribution of each structural prior to the generated virtual points or soft constraints. The entire system is trained end-to-end via backpropagation, with gradient signals propagating through both the structural modules and the selection mechanism.

This design provides several advantages. First, it improves sample efficiency by encouraging inductive biases that reduce overfitting and help the learner generalise across novel or abstract tasks. Second, it enables the emergence of useful structural generalisations such as rotational invariance or compositional semantics without requiring explicit supervision. Third, the interpretability of algebraic or geometric structures offers insight into what the system has learned. Lastly, the ability to compose retained structures across meta-levels facilitates abstraction and curriculum learning over increasingly general problem families.

As an illustrative example, consider a generator that initially encodes both rotation (\( SO(2) \)) and translation (\( \mathbb{R}^2 \)) symmetries. For a class of robotic manipulation tasks, the system may learn that only rotational invariance contributes consistently to generalisation. Consequently, the selector assigns near-zero weight to the translation module, effectively pruning it from the constraint manifold for that task distribution.


\paragraph{Improving Sample Efficiency in Recursive Meta-Learning.}
To enhance the sample efficiency of the general recursive meta-learning framework, we introduce several architecture-agnostic mechanisms applicable across different levels of abstraction and task domains. Firstly, the reuse of previously collected data is enabled through off-policy replay or task buffer sampling, allowing for more data-efficient updates without resampling from the environment. Secondly, surrogate loss functions are learned to approximate the true task losses, enabling gradient-based updates without full inner-loop optimisation, thereby reducing computational cost. Thirdly, amortised inference mechanisms are introduced at each level to initialise lower-level learners via learned mappings conditioned on task embeddings, minimising warm-start cost. Fourth, active meta-task selection based on learning progress or epistemic uncertainty allows the meta-learner to focus on informative regions of the task space. Fifth, synthetic data generation via generative models (e.g., VAEs~\cite{pin2021} or GANs~\cite{goodfellow2020generative}) facilitates virtual task sampling and regularisation, extending generalisation while avoiding expensive real task instantiations. Sixth, unrolled inner-loop learning is backpropagated through to enable meta-level adaptation to task dynamics. Finally, cross-level conditioning allows higher-level learners to integrate representations and parameters of their lower-level counterparts, promoting abstraction and structured knowledge transfer throughout the recursive hierarchy.

An example surrogate loss function used in a supervised setting may take the form:
\[
\mathcal{L}^{\Phi_k}_{\text{surrogate}}(f_\theta, \tau) = \|f_\theta(x_\tau) - y_\tau\|^2 + \lambda \|\nabla_\theta f_\theta(x_\tau)\|^2,
\]
where \( (x_\tau, y_\tau) \) are sampled from \( \tau \), and the second term penalises sharp loss landscapes, encouraging smoother adaptation trajectories. Integrating these efficiency modifications leads to our Sample-Efficient Recursive Meta-Learning algorithm (Algorithm \ref{alg:sample_efficient_meta_learning}).

\section{Category-Theoretic Perspective}
\label{sec:category_th}
We now give an overview of the conceptual structure of our framework from the category-theoretic perspective and define the relevant objects.  Category theory provides a unifying mathematical language for representing the structures and transformations present in hierarchical meta-learning. A \textit{category} $\mathcal{C}$ consists of a collection of \textit{objects}, which may represent mathematical structures such as data sets, models, or tasks, together with \textit{morphisms} (arrows) between these objects, which represent structure-preserving transformations. Morphisms can be composed associatively, and each object possesses an identity morphism. A \textit{functor} $F : \mathcal{C} \rightarrow \mathcal{D}$ maps objects in category $\mathcal{C}$ to objects in category $\mathcal{D}$, and morphisms to morphisms, preserving the composition and identity structure of the original category. A \textit{natural transformation} relates two functors between the same categories, capturing the idea of a systematic transformation of mappings between entire families of objects.

In our hierarchical meta-learning framework, these categorical constructs provide a formal abstraction of the learning process. The set of lower-level tasks can be modelled as a category $\mathcal{T}$, whose objects are task instances (e.g., specific games, data sets, or environments), and whose morphisms represent task transformations or equivalences (such as symmetry operations or data augmentations). The base learner acts as a \textit{functor} $F_0 : \mathcal{T} \rightarrow \mathcal{S}$, where $\mathcal{S}$ is a category of solutions, such as trained models. The task learner or meta-learner at level 1 constructs transformations or constraints that systematically shape the learning performed by $F_0$. Conceptually, this corresponds to a higher-order functor $F_1 : \mathcal{T} \rightarrow \text{Fun}(\mathcal{T}, \mathcal{S})$, where $\text{Fun}(\mathcal{T}, \mathcal{S})$ is the functor category whose objects are functors (i.e., base learners) and whose morphisms are natural transformations between learners. Therefore, the level 1 meta-learner learns to generate soft constraints and task transformations that guide $F_0$. In turn, the level 2 meta-learner can be viewed as a functor $F_2 : \mathcal{T} \rightarrow \text{Fun}(\mathcal{T}, \text{Fun}(\mathcal{T}, \mathcal{S}))$, learning to generate learning rules for meta-learners. In this view, the recursive architecture builds a tower of functors and natural transformations, each abstracting the learning dynamics of the layer below.

This categorical perspective allows us to capture the structure of the models and tasks and, the compositionality and generality of the learning process. Meta-learning corresponds to learning functorial mappings between categories of problems and solution spaces, with higher-order meta-learning operating on categories of learning processes. Collecting the above objects yields the following description given a cateory of tasks \( \mathcal{T} \)
\begin{itemize}
    \item Objects: learning tasks \( T \).
    \item Morphisms: task transformations (e.g., domain shifts, augmentations).
\end{itemize}

For a  \textit{category of models}  \( \mathcal{M} \):
\begin{itemize}
    \item Objects: parameterised models \( M \).
    \item Morphisms: model mappings (e.g., fine-tuning, parameter transfers).
\end{itemize}

\noindent*\textit{Learner:} A functor: $
L : \mathcal{T} \to \mathcal{M}$. \newline*\textit{Meta-learner:} A natural transformation: $
\eta: L_1 \Rightarrow L_2$. \newline*\textit{Higher-order meta-learner:} A functor on functor categories:
$
F_k : \text{Fun}(\mathcal{T}, \mathcal{M}) \to \text{Fun}(\mathcal{T}, \mathcal{M})$.

This recursion allows us to stack learners of learners, in a modular and theoretically sound way. From a category-theoretic perspective, this hierarchical meta-learning framework arises naturally through the composition of functors. Each learner, meta-learner, and higher-order meta-learner can be formalised as a functor between appropriate categories of models and tasks. The base learner can be viewed as a functor \( L : \mathcal{T} \to \mathcal{M} \), mapping tasks to models. The meta-learner is a functor \( F_1 : \mathcal{T} \to \mathcal{L} \), where \( \mathcal{L} \) is the category of learning algorithms or model-generating procedures. The meta-meta-learner is a functor \( F_2 : \mathcal{D} \to \mathcal{F}_1 \), where \( \mathcal{D} \) is the category of domains and \( \mathcal{F}_1 \) is the category of meta-learning processes. This layered composition of functors elegantly captures the notion that learning strategies themselves can be learned, and that higher-order regularities about learning can be discovered through further abstraction. Moreover, the categorical view provides formal mathematical tools for understanding the compositionality, reusability, and transferability of learning processes across levels of abstraction, enabling principled design and analysis of multi-level meta-learning systems.

Consider a category \(\mathcal{L}\) whose objects are \emph{learners}, i.e., parametrised models or algorithms that map data sets to hypotheses. Morphisms in \(\mathcal{L}\) are learning transformations or adaptations between learners, such as updating parameters or transferring knowledge. Each meta-learning level induces an endofunctor\footnote{An endofunctor is a functor that maps a category to itself, assigning to each object and arrow in the category another object and arrow in the same category, while preserving identities and composition.}
\[
F_k : \mathcal{L} \to \mathcal{L},
\]
mapping a learner \(L_{k-1}\) at level \(k-1\) to a meta-learner \(L_k = F_k(L_{k-1})\) at level \(k\). The composition of these functors
\[
F_K \circ F_{K-1} \circ \cdots \circ F_1,
\]
models the full recursive hierarchy, representing a multi-level adaptation pipeline. 


%

Our framework admits a natural categorical interpretation that clarifies its compositional and recursive nature. At each meta-level \(k\), we view the meta-learner \(L_k\) as a functor between categories of learning tasks and models. Concretely, define by \(\mathbf{Task}_k\), the category whose objects are tasks \(T_k\) at level \(k\), and morphisms represent task transformations or generalisations and by \(\mathbf{Model}_k.\), the category of parametrised models or learners \(f_{\theta_k}\) associated to \(\mathbf{Task}_k\).
Then the meta-learner \(L_k : \mathbf{Task}_{k-1} \to \mathbf{Model}_{k-1}\) induces a mapping at the level of categories, satisfying functoriality conditions that preserve task compositions and model adaptations. This structure extends recursively:

\[
L_K \circ L_{K-1} \circ \cdots \circ L_1 : \mathbf{Task}_0 \to \mathbf{Model}_0,
\]
expressing the entire hierarchy as a composite functor.

This categorical viewpoint offers several benefits. \textit{Modularity and Compositionality:} each meta-level can be studied and optimised independently, then composed, respecting structural relationships. \textit{Fixed-Point Semantics:} recursive meta-learning corresponds to finding fixed points of functorial transformations, a well-studied concept in category theory.  \textit{Abstraction and Generalisation:} morphisms in \(\mathbf{Task}_k\) capture abstractions and curriculum relations, enabling formal reasoning about knowledge transfer across levels. This framework also aids in formalising the inter-level constraints (soft constraints) as natural transformations, bridging different meta-level learners coherently.

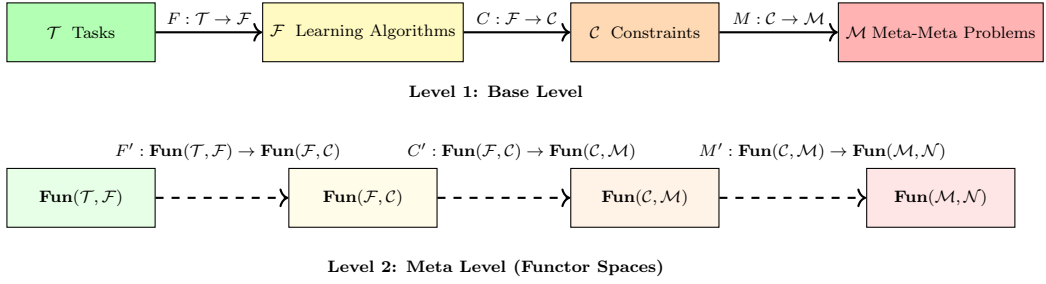
\begin{figure}[t]
\begin{tikzpicture}[node distance=1.8cm and 1.8cm, every node/.style={font=\scriptsize}, scale=0.78, transform shape]

\node (T) at (0,0) [draw, rectangle, minimum width=2.5cm, minimum height=1cm, fill=green!30] {\textbf{$\mathcal{T}$} \\\; Tasks};
\node (F) [draw, rectangle, minimum width=2.5cm, minimum height=1cm, fill=yellow!30, right=of T] {\textbf{$\mathcal{F}$} \\\; Learning Algorithms};
\node (C) [draw, rectangle, minimum width=2.5cm, minimum height=1cm, fill=orange!30, right=of F] {\textbf{$\mathcal{C}$} \\\; Constraints};
\node (M) [draw, rectangle, minimum width=2.5cm, minimum height=1cm, fill=red!30, right=of C, xshift=.2cm] {\textbf{$\mathcal{M}$} \\\;Meta-Meta Problems};

\draw[->, thick] (T) -- node[above] {$F: \mathcal{T} \to \mathcal{F}$} (F);
\draw[->, thick] (F) -- node[above] {$C: \mathcal{F} \to \mathcal{C}$} (C);
\draw[->, thick] (C) -- node[above] {$M: \mathcal{C} \to \mathcal{M}$} (M);

\node (TF) [draw, rectangle, minimum width=2.5cm, minimum height=1cm, fill=green!10, below=of T] {\textbf{$\mathbf{Fun}(\mathcal{T}, \mathcal{F})$}};
\node (FF) [draw, rectangle, minimum width=2.5cm, minimum height=1cm, fill=yellow!10, below=of F] {\textbf{$\mathbf{Fun}(\mathcal{F}, \mathcal{C})$}};
\node (CF) [draw, rectangle, minimum width=2.5cm, minimum height=1cm, fill=orange!10, below=of C] {\textbf{$\mathbf{Fun}(\mathcal{C}, \mathcal{M})$}};
\node (MF) [draw, rectangle, minimum width=2.5cm, minimum height=1cm, fill=red!10, below=of M] {\textbf{$\mathbf{Fun}(\mathcal{M}, \mathcal{N})$}};

\draw[->, dashed, thick] (TF) -- node[above, xshift=.1cm, yshift=.5cm] {$F': \mathbf{Fun}(\mathcal{T}, \mathcal{F}) \to \mathbf{Fun}(\mathcal{F}, \mathcal{C})$} (FF);
\draw[->, dashed, thick] (FF) -- node[above,xshift=.3cm, yshift=.5cm] {$C': \mathbf{Fun}(\mathcal{F}, \mathcal{C}) \to \mathbf{Fun}(\mathcal{C}, \mathcal{M})$} (CF);
\draw[->, dashed, thick] (CF) -- node[above,xshift=.5cm, yshift=.5cm] {$M': \mathbf{Fun}(\mathcal{C}, \mathcal{M}) \to \mathbf{Fun}(\mathcal{M}, \mathcal{N})$} (MF);

\node at (7,-1) {\textbf{Level 1: Base Level}};
\node at (7,-4.0) {\textbf{Level 2: Meta Level (Functor Spaces)}};
\end{tikzpicture}
\caption{\small  Categorical view of the hierarchical meta-learning framework as a multi-level functor stack. Each level corresponds to functors between categories representing tasks, learning algorithms, constraints, and meta-problems. Dashed arrows indicate meta-meta functors mapping between functor spaces themselves.}
\label{fig:functor-stack}
\end{figure}

The \emph{Yoneda Lemma} is one of the foundational results in category theory, with deep implications for the nature of objects and their representations within a categorical framework~\cite{kudasov2024formalizing,rasekh2023yoneda}. The Yoneda Lemma states that for any such functor $F: \mathcal{C}^{\mathrm{op}} \rightarrow \mathbf{Set}$, and any object $A$ in $\mathcal{C}$, there is a natural bijection\footnote{To state it formally, let $\mathcal{C}$ be a locally small category, meaning that for any two objects in $\mathcal{C}$, the collection of morphisms between them forms a set. For any object $A$ in $\mathcal{C}$, one may define a functor $h_A := \mathrm{Hom}_\mathcal{C}(-, A)$ from $\mathcal{C}^{\mathrm{op}}$ (the opposite category of $\mathcal{C}$) to the category of sets, $\mathbf{Set}$. This functor assigns to each object $X$ the set of morphisms from $X$ to $A$, and to each morphism $f: X \rightarrow Y$ in $\mathcal{C}$, the function that maps $\phi \in \mathrm{Hom}_\mathcal{C}(Y, A)$ to $\phi \circ f \in \mathrm{Hom}_\mathcal{C}(X, A)$.
}
\[
\mathrm{Nat}(h_A, F) \cong F(A),
\]
where $\mathrm{Nat}(h_A, F)$ denotes the set of natural transformations from $h_A$ to $F$. This bijection is natural in both $A$ and $F$, meaning it respects the structure of the category and the behaviour of functors. Intuitively, the lemma asserts that the identity of an object is fully captured by the pattern of morphisms into it: that is, an object can be uniquely understood in terms of how it is seen by all other objects in the category.

In the context of recursive meta-learning, we may apply this principle to the structure of learners and meta-learners. We model the domain of tasks as a category $\mathcal{T}$; learning algorithms as objects of a category $\mathcal{F}$; and learning strategies (or learning constraints) as morphisms within $\mathcal{F}$. A learner is then a functor $F: \mathcal{T} \rightarrow \mathcal{F}$ that maps each task to a corresponding algorithm, preserving task structure in algorithmic space. The meta-learner is a transformation between such functors (and not just a single functor). Specifically, a meta-learner is modelled as a natural transformation $\eta: G \Rightarrow F$, which consists of a collection of morphisms $\eta_T: G(T) \rightarrow F(T)$ in $\mathcal{F}$, one for each task $T$, such that these morphisms commute with the task structure encoded in $\mathcal{T}$.

Yoneda's Lemma now offers a profound insight--- it tells us that to understand a given learner $F$, it is sufficient to understand how it relates to all other learners via natural transformations. That is, its identity is encoded entirely by the set of such natural transformations. For meta-learning, this means that the task of discovering or optimising learners can be formulated in terms of the morphisms between functors i.e., through the relationships between different learners on the entire space of tasks. Consequently, the space of learners forms a functor category $\mathbf{Fun}(\mathcal{T}, \mathcal{F})$, and the higher-order learning carried out by meta-learners resides in the space of morphisms within this category.

In Section \ref{subsec:exploration_sec}, we constructed a strategy to explore the virtual task space in which the meta-learner uses a generative model that seeks out points in the virtual point manifold that maximise the lower level's expected difficulty. From a categorical perspective, the virtual task generator can be viewed as constructing new morphisms in the functor categories or higher, thereby enriching the morphism space with transformations that induce learning dynamics. In particular, categorically, this process can be understood as enriching the space of morphisms in the functor categories $
F': \mathbf{Fun}(\mathcal{T}, \mathcal{F}) \to \mathbf{Fun}(\mathcal{F}, \mathcal{C})$, 
$F'$ captures the transformation induced by the virtual task generator. Each $\tilde{x} = G_\phi(z)$ corresponds to a morphism revealing the learner's functional behaviour across tasks.

The adversarial exploration can be interpreted categorically. Each level of the framework operates within a functor category $\mathbf{Fun}(\mathcal{C}_k, \mathcal{C}_{k-1})$, where objects are spaces of theorems and morphisms represent proof-preserving transformations. The virtual point generator samples candidate morphisms (i.e., transformations between theorem instances) that are unlikely to be inhabited by successful proofs given the current learner. 

 Yoneda's lemma implies that to understand a task object, it suffices to examine how it is mapped to and from other objects via such morphisms. 
 The learner's interaction with these virtual morphisms allows the meta-learner to reconstruct, approximate, and eventually generalise the constraint via observed responses which is precisely in the spirit of Yoneda's Lemma.
 In this light, the meta-learner's exploration can be seen as learning to construct virtual points that reveal the full representational role of task types, abstracting over instantiations to uncover the structure-preserving maps that define generalisable knowledge.
 Consequently, the exploration strategy acts as an empirical probe of this functorial behaviour --- only those abstract theorems whose morphisms are consistently realisable across sampled virtual problems can be retained as valid generalisations. As virtual points evolve through the recursive levels, they encode increasingly abstract relationships, potentially corresponding to whole task types or regimes (e.g., classes of games or differential equations) rather than specific instances. 


A particularly compelling application of Yoneda’s Lemma within our recursive meta-learning framework arises in the context of understanding highly complex learners whose internal specifications are too intricate to model directly. Rather than relying on explicit knowledge of a model's internal structure, we could model learners via the pattern of their behavioural responses across a systematically constructed space of tasks. This aligns naturally with our recursive meta-learning setup, where higher-level meta-learners are tasked with uncovering abstract generalisations and regularities from the observed behaviours of lower-level learners. The human brain stands as an extreme example: its architecture and dynamics elude precise mechanistic reconstruction, yet its learning behaviour can be probed through its responses by functorially representing it in terms of its external performance profile. Importantly, this framework allows for structured and scalable exploration, potentially mitigating the need for costly or impractical training within real-world environments. 

\section{Curriculum Learning}

The hierarchical meta-learning framework described here can be naturally interpreted as a form of \emph{curriculum learning}~\cite{bengio2009curriculum,hacohen2019power} in which each level corresponds to a generalised abstraction of the problem space faced by the level below. In traditional curriculum learning, models are presented with training examples ordered from simple to complex, gradually building their capacity to solve harder problems. In our framework, this progression occurs \emph{along the abstraction axis}: each higher meta-level is exposed not to increasingly complex individual tasks, but to increasingly abstract \emph{classes of tasks}, learning to generalise inductive biases, learning strategies, and meta-strategies across broader conceptual spaces.

\normalsize
At the base level, a learner solves individual tasks, such as fitting a function or classifying images. The meta-learner abstracts over individual tasks within a domain, learning how to produce learners that adapt quickly and generalise well. The meta-meta-learner abstracts over \emph{families of domains}, learning relationships between domain properties and optimal learning behaviours. In this view, the full stack of learners and meta-learners constitutes an \emph{abstraction curriculum}, where each level internalises and generalises the lessons of the previous one.

Therefore, this multi-level framework can be viewed as a form of curriculum learning in which the curriculum is not ordered over individual examples, but over \emph{conceptual abstractions} and \emph{meta-relationships}. Each level learns the ``lessons'' of the level below and abstracts them, progressively building higher-order knowledge about learning itself. The category-theoretic formulation supports this view by treating each meta-level as a functor that generalises the mappings of the level below. The entire stack therefore forms an abstraction pipeline that can be composed and analysed formally.

\[
\begin{array}{lcl}
\textbf{Level 2: Meta-Meta-Learner} & \Rightarrow & \text{Abstracts over Families of Domains} \\
\Downarrow & & \text{(e.g.\ different physical, biological, control systems)} \\
\textbf{Level 1: Meta-Learner} & \Rightarrow & \text{Abstracts over Tasks within Domain} \\
\Downarrow & & \text{(e.g.\ different ODE instances within a system)} \\
\textbf{Level 0: Learner} & \Rightarrow & \text{Solves Specific Task Instance} \\
\Downarrow & & \text{(e.g.\ solve ODE with given \( \theta \))} \\
\end{array}
\]

\noindent This abstraction curriculum enables progressive generalisation across increasingly abstract levels of learning, with each meta-level guiding and structuring the learning process below.

\begin{center}
\fbox{%
\parbox{0.95\textwidth}{%
\textbf{Curriculum Along Abstraction Axis.} The hierarchical meta-learning framework constitutes a form of curriculum learning in which each meta-level abstracts over the level below. The base learner solves specific instances; the meta-learner generalises across tasks within a domain; the meta-meta-learner generalises across families of domains. This constitutes an \emph{abstraction curriculum}, whereby progressively more abstract regularities about learning are acquired and transferred across levels. Our framework therefore enables principled and scalable generalisation of inductive biases, learning strategies, and meta-strategies.
}%
}
\end{center}




\paragraph{Computational Considerations.}

While the hierarchical meta-learning framework provides strong conceptual advantages, it also introduces significant computational considerations. Each additional meta-level requires the optimisation of increasingly abstract objectives, typically involving higher-order gradient computations and nested loops. This can lead to increased training times and memory usage, particularly when virtual tasks and soft constraints are used extensively. However, this cost is offset by practical benefits: improved transfer learning across domains, more robust inductive biases, and the ability to generate models that generalise well even in low-data regimes. Techniques such as first-order meta-gradient approximations, task subsampling, and memory-efficient implicit differentiation can help mitigate computational costs in practice.

\begin{tcolorbox}[%
    enhanced, 
    breakable,
    frame hidden,
    overlay broken = {
        \draw[line width=.4mm, black, rounded corners]
        (frame.north west) rectangle (frame.south east);},
    ]{}\subsection*{Example 2: Solving Parametrised Families of Differential Equations}
\small
Consider the problem of learning to solve parametrised families of differential equations. At the lowest level, the learner solves the individual ordinary differential equations (ODEs) $
\frac{dy}{dt} = f_{\theta}(y, t)$,
for a fixed parameter \( \theta \). The task at this level is to approximate the solution \( y(t) \) for a given \( \theta \) and initial condition.

At the meta-learner level, the system learns to produce learners that can quickly adapt to new instances of the ODE with different parameters \( \theta \), given few examples of \( y(t) \). For instance, the meta-learner might discover that certain solution structures (such as exponential decay, oscillatory dynamics, or stable manifolds) frequently arise, and therefore bias base learners toward these forms through soft constraints or inductive priors. At the meta-meta-learner level, the system is trained across \emph{different families of differential equations}, for example, physical systems (Hamiltonian or Lagrangian dynamics), biological systems (population dynamics), control systems (feedback-controlled ordinary differential equations (ODEs)). 
The meta-meta-learner learns abstract relationships between the \emph{family of equations} and the meta-learning strategy that should be applied. It may learn, for example, that Hamiltonian systems should preserve symplectic structure and energy constraints, and therefore adjust the meta-learner to inject appropriate inductive biases (via soft constraints) when faced with such domains.

In this way, the system progresses through an \emph{abstraction curriculum}
\begin{itemize}
    \item Level 0: Solve specific ODE with given \( \theta \)
    \item Level 1: Learn to quickly adapt to new \( \theta \) for the same ODE family
    \item Level 2: Learn how different families of ODEs relate to learning biases and strategies
\end{itemize}

At each level, what was a variable of interest at the lower level becomes part of the \emph{contextual meta-structure} that the higher level generalises over. This demonstrates how hierarchical meta-learning can be seen as a natural extension of curriculum learning along the axis of abstraction and generalisation.
\end{tcolorbox}

\section{Abstraction in Higher-Order Meta-Learning}
\label{sec:abstraction}
Higher-order meta-learning provides a powerful mechanism for enabling learning systems to acquire \emph{abstract generalisations} about the learning process itself. At each level of our meta-learning stack, the scope of what is being learned progressively expands, allowing the system to discover higher-order patterns that transcend individual tasks and domains. Our architecture which is illustrated in Figure~\ref{fig:architecure_2} can be understood as a form of curriculum learning along the axis of abstraction. Rather than ordering training examples from simple to complex within a fixed problem space, our framework organises learning hierarchically across conceptual levels. Each meta-level abstracts and generalises over the learning dynamics of the level below. This enables the system to acquire progressively more abstract inductive biases and meta-strategies, improving its ability to transfer knowledge across domains and families of domains. The category-theoretic formulation provides a principled way to model and analyse this compositional structure.

The specifics are as follows: at the base level, a learner is trained to produce solutions to specific tasks. A first-level meta-learner, however, is not concerned with specific task solutions, but with learning \emph{how to produce learners} that generalise well across a family of related tasks. In the same example, it might learn an inductive bias or initialisation that enables rapid adaptation.

As we ascend the hierarchy, a second-level meta-learner, or meta-meta-learner, is introduced. Its role is to discover \emph{abstract patterns governing the effectiveness of different meta-learning strategies} across broader and more diverse families of tasks. Crucially, this enables the system to recognise relationships between the structure of task families and the meta-learning behaviours that are most effective for them. In this sense, the meta-meta-learner does not merely adapt models to tasks; it adapts \emph{meta-learning processes} to families of domains.
%
Through this process, the meta-meta-learner acquires \emph{abstract generalisations about learning itself}. It learns that domains with high intra-class variability benefit from robust, invariant representations; that texture-sensitive domains should emphasise local pattern filters; and that geometric transformations require appropriate augmentation strategies and equivariance constraints. These insights are not specific to any single task but capture general relationships between the nature of a domain and the optimal form of meta-learning to apply.

This framework therefore enables a learning system to develop a rich hierarchy of knowledge: learning how to solve tasks, how to learn to solve tasks, and how to learn how to learn, adapting its own learning processes in a principled, data-driven manner. The recursive structure provided by category theory ensures that these layers of learning are coherently composed, while the soft constraints and virtual tasks introduced at each level allow for the flexible enforcement of desirable properties and inductive biases.


An intriguing property of the proposed hierarchical meta-learning framework is that each meta-learner actively explores the space of learning rules, constraints, and inductive biases that govern subordinate-level learning. The meta-learner operates at the level of shaping learning \emph{processes}, rather than directly mapping data to outputs.

Exploration at the meta-level can proceed through several mechanisms. Firstly, the meta-learner generates virtual tasks by sampling within or extrapolating from the space of observed tasks. Secondly, it explores the parameter space of soft constraints, parameterised as neural networks, symbolic logic, or energy functions through gradient-based meta-optimisation. Thirdly, stochastic exploration strategies, such as adversarial generation of virtual tasks or random perturbations of constraints, drive discovery of useful inductive structures. Moreover, the meta-learner may optimise principled objectives, for example PAC-Bayes generalisation bounds over subordinate learners, to guide exploration toward robust, generalisable solutions.

\subsection{Discovering Behavioural Equivalences across Learners}

We now turn to a case in which the learners themselves are the object of focus. In this instance, we look to patterns across different low level learners' behaviour given some fixed task with the aim of optimising the learner processes.  Let $\mathcal{L}_\omega$ denote a learner indexed by a parameter $\omega \in \Omega$, where $\Omega$ is a space of algorithmic, architectural, or structural representations (e.g., Q-learning~\cite{watkins1992q}, PPO~\cite{schulman2017proximal}, LSTMs~\cite{graves2012long}, or differentiable solvers). In many domains, especially in reinforcement learning and game theory, we observe scenarios in which different learners solve the same task $\tau$ with indistinguishable performance, or conversely, where similar learners yield highly divergent outcomes. Our framework provides a systematic way to identify, compare, and generalise such behaviours across structured spaces of learners. Suppose $\omega_1, \omega_2 \in \Omega$ such that the corresponding learners $\mathcal{L}_{\omega_1}$ and $\mathcal{L}_{\omega_2}$ are structurally dissimilar (e.g., differing update rules or internal representations), but exhibit similar performance on a given task $\tau$, i.e.,
\[
\mathcal{R}(\mathcal{L}_{\omega_1}, \tau) \approx \mathcal{R}(\mathcal{L}_{\omega_2}, \tau).
\]
The higher levels of the meta-learning hierarchy may learn to compress or categorise such learners into equivalence classes, organised not by internal mechanisms but by externally observed behaviour. This reveals deep regularities, uncovers task-invariant transformations, and supports learner generalisation through behavioural embedding. These equivalence relations can be learned directly from trajectories, outcomes, or loss landscapes, allowing the meta-learner to infer structure even when learners are black-box.

Conversely, when $\omega_1 \approx \omega_2$ but
\[
\mathcal{R}(\mathcal{L}_{\omega_1}, \tau) \gg \mathcal{R}(\mathcal{L}_{\omega_2}, \tau),
\]
the meta-learner can detect sharp performance bifurcations. Such situations indicate sensitivity to minor variations in model or optimisation design. The framework can identify such failure regions and adapt constraints or generate virtual tasks to probe them more deeply. This supports the discovery of structurally fragile learners, automatically learned regularisation terms, and useful inductive biases.

At level $0$, learners execute their task-specific optimisation. At level $1$, meta-learners generalise over performance across parametrised learners, learning to predict behaviour or performance across $\Omega$. At level $2$ and beyond, learners generalise across entire families of $\Omega$-spaces, capturing abstract patterns of equivalence, divergence, and symmetry. These higher-level abstractions allow the framework to automatically build representations of functional classes, generalisable update rules, and even task-specific solver design.

In reinforcement learning, learners such as DQN, PPO, or actor-critic methods may differ in architecture, gradient structure, or inductive priors, yet converge to similar behaviours on a specific environment. The recursive meta-learner can identify which structural choices are functionally equivalent, predict which policies generalise better, or discover new policy update objectives. It may also identify fragile algorithm-task pairings or synthesise virtual environments that discriminate between subtly different learners.

In algorithmic game theory, different solvers (e.g., fictitious play, best response, regret minimisation) often converge in coordination or team games but diverge in general-sum games. The meta-learner can quantify this divergence across game classes and use its recursive structure to infer general properties required for convergence (e.g., monotonicity, potential structure, no-regret dynamics). This can guide the synthesis of new algorithms or refine solver selection based on game-theoretic structure.

The ideas above are conceptually aligned with the insight from Yoneda’s lemma: objects (learners) are characterised by their behaviour in all contexts (tasks) rather than by their internal constitution. Behaviourally equivalent learners cannot be distinguished through their observed responses to tasks, and the meta-learner discovers this via task-induced exploration. This underscores the power of probing and observation as a means of understanding highly complex or opaque learners.

\begin{tcolorbox}[%
    enhanced, 
    breakable,
    frame hidden,
    overlay broken = {
        \draw[line width=.4mm, black, rounded corners]
        (frame.north west) rectangle (frame.south east);},
    ]{}\subsection*{Example 3: Learning Nash Equilibria in Games}
\label{sec:game_th} \small





Our framework offers a natural fit for problems involving structured convergence and solution inference in game theory i.e. for computing \textit{Nash equilibria}~\cite{osborne1994course} across families of normal form games. We demonstrate how the recursive framework can be instantiated to determine the required properties for convergence to Nash equilibria across structured game classes and more generally learning equilibria within normal form games.
Computing Nash equilibria is computationally hard in the general case~\cite{10.1145/1516512.1516516}. However, several subclasses of games admit more structure and tractability. Three such examples are \textit{Team games}~\cite{monderer1996fictitious} are games in which the equilibrium reduces to jointly maximising the shared utility. \textit{Coordination games}~\cite{FARRELL1988209} in which players are rewarded for selecting aligned strategies  and \textit{Potential games}~\cite{MONDERER1996124} for which there exists a function $\Phi$ known as the potential function which quantifies the change in reward for a player given a unilateral deviation by that player.
The three classes of games exhibit a structured hierarchy: Team Games $\subset$ Coordination Games $\subset$ Potential Games.\footnote{In a potential game each player's utility function can be expressed as \( u_i(a_1,\ldots, a_n) = \Phi(a_1,\ldots, a_n) + F_i(a^{-i}) \), where \( F_i \) depends only on the opponents' actions' $a^{-i}\in A_{-i}$.  Writing \( u_i = \phi + \lambda F_i \) enables a smooth interpolation between team games and potential games.} Therefore, computing the Nash equilibria across these classes represents learning with increasing  generality and complexity. Within this hierarchy, algorithms that converge reliably in simpler settings such as consensus methods in team games may exhibit instability or divergence when applied to more general classes, such as potential games or general-sum games~\cite{balduzzi2018mechanics}.

We may consider a stratified learning process across three levels. At the base level, the learner is trained on team games---games in which both players receive identical payoffs. These games are structurally simpler, as the interests of the players are perfectly aligned, and so the equilibrium concept reduces to joint maximisation. At the next level, the learner is exposed to coordination games, in which multiple Nash equilibria may exist, but there is a shared incentive to coordinate behaviour. These are more complex than team games, but retain a degree of alignment in objectives. Finally, at a higher level still, the learner is trained on potential games. These games possess a potential function whose local maxima correspond to Nash equilibria, but the players' payoffs are not necessarily even similar. Each class of game corresponds to a subcategory of $\mathcal{T}$, and the learning process requires increasing generalisation as we ascend this hierarchy. The recursive meta-learning framework offers a structured and systematic approach to uncovering the conditions under which learning algorithms converge across different classes of tasks. 

In this context, the meta-learning framework enables both prediction and modification of convergence behaviour. As we ascend the hierarchy, meta-learners are exposed to increasingly general game classes eventually being exposed to potential games or non-potential general-sum games. These learners can identify structural regularities in the space of task-response behaviours, and can thereby learn which learning algorithms or update rules fail or succeed in particular settings.

The virtual task generation serves to probe transition points; regions in task space where convergence behaviour changes qualitatively. Crucially, higher-level insights allows the framework to guide exploration toward regions of the task space that are more pedagogically informative. A higher-level learner, may synthesise virtual tasks not by replicating previously encountered games, but by generating instances that instantiate broader principles of convergence. For instance, by generating games that interpolate between a coordination game and a general-sum game, the meta-learner can observe when a given base-level algorithm (e.g., Q-learning) begins to exhibit instability. These virtual points can therefore serve as exemplars of more abstract game structures, implicitly nudging subordinate learners toward internalising the inductive biases necessary for generalisation across game classes. This supports the data-driven discovery of convergence-preserving structures, such as alignment between gradient directions and payoff landscapes, or the preservation of monotonicity in best-response dynamics.


From a category-theoretic standpoint, the framework models learners as functors between task categories and behavioural categories, preserving structural relationships between tasks and their induced learning dynamics. The higher-order functors at the meta-level, in turn, model transformations between learners, enabling the abstraction of update rules and regularisation principles. These abstractions form the basis for meta-learned inductive biases that govern convergence more broadly, even in the absence of direct analytical guarantees. Yoneda's Lemma further justifies the ability of higher-level meta-learners to infer the convergence properties of base-level learners purely from behavioural observations across a wide and well-chosen range of virtual tasks, even without explicit access to the internal details of the learning algorithm. Thus, the lemma provides a theoretical foundation for probing complex learning dynamics through their external manifestations in a structured and sample-efficient way.
\\\textbf{Curriculum learning perspective.} We can also consider our framework from the perspective of curriculum learning. Our category-theoretic insights become especially useful in this setting. Consider the case where $\mathcal{T}$ is the category of such games, with morphisms given by appropriate structure-preserving transformations (such as game embeddings or homomorphisms), and $\mathcal{F}$ is the category of learning procedures or solution concepts. In this setting, a learner $F: \mathcal{T} \rightarrow \mathcal{F}$ assigns to each game a method for computing its equilibria. The recursive meta-learning framework provides a powerful language for understanding how structured game-theoretic curricula might emerge endogenously. Each category $\mathcal{T}$, $\mathcal{F}$, $\mathcal{C}$, and $\mathcal{M}$ in the hierarchy encapsulates a level of abstraction ranging from concrete games to learning strategies, constraints, and finally meta-constraints governing learning processes themselves. In this context, the system does not require a priori specification of whether a particular game belongs to a known class such as team, coordination, or potential games. Instead, each game is treated as an object within a category of tasks, and transformations between these tasks such as the learning algorithms that infer equilibria, are captured by morphisms. As the learner ascends the hierarchy, functors map entire structures of objects and morphisms from one category to the next, enabling increasingly abstract representations of learning and strategic reasoning.

\end{tcolorbox}

Beyond exploration, this framework facilitates the emergence of new meta-problems and functor spaces. At each level, the learner discovers mappings from tasks to solutions or from problems to learning procedures. From a categorical perspective, this corresponds to the discovery of functors between categories of tasks and categories of learned models or procedures. Higher-level meta-learners analyse patterns in these subordinate-level functors, identifying regularities, symmetries, and compositional structures.

For instance, clustering subordinate-level functors based on induced learning dynamics reveals latent problem space structure. The meta-learner may discover families of learning rules that generalise across distinct domains, prompting the formulation of higher-order meta-problems: learning functors between functor spaces themselves. Meta-composition, the combination of learned functors drives the generation of novel composite learning strategies and the emergence of previously unconsidered task structures.

Such capabilities suggest the potential for open-ended learning systems that autonomously invent new problems to solve and new methods to solve them. In principle, our framework could enable neural networks not only to master a fixed distribution of tasks but also to continuously expand their own problem space and curriculum, achieving artificial creativity and self-directed learning. The categorical formulation provides a formal language for such hierarchical abstraction, composition, and emergence.

\begin{tcolorbox}[%
    enhanced, 
    breakable,
    frame hidden,
    overlay broken = {
        \draw[line width=.4mm, black, rounded corners]
        (frame.north west) rectangle (frame.south east);},
    ]{}\subsection*{Example 4: Learning Policy Update Objectives}\label{example:learning_RL_updates} \small

A fundamental question in reinforcement learning (RL) is how to design effective policy update rules that lead to robust and generalisable behaviours across diverse environments. Traditionally, update rules such as policy gradients or actor-critic methods are specified manually, drawing on theoretical insights or empirical heuristics. However, these hand-crafted objectives may not be optimal across different problem domains or training regimes. In this example, we give an instantiation of our recursive meta-learning framework designed to autonomously discover effective policy update objectives. This is achieved through a nested learning procedure in which both the form and parameters of update rules are learned from data, potentially outperforming fixed hand-designed strategies.

Our approach is organised into a hierarchical structure. At the base level (level 0), a standard RL agent trains a policy using an update rule whose functional form is determined by a meta-learner operating at level 1. This meta-learner is responsible for learning the structure of the policy update loss function by observing performance across a distribution of RL tasks. At level 2, a meta-meta-learner generalises across these meta-learning processes, potentially learning priors or constraints over the space of loss functions and generating synthetic tasks or training signals to guide generalisation.

At the core of this procedure is the idea that both the structure and parameters of the policy update loss function are represented as neural networks and optimised via backpropagation through the entire learning trajectory. This requires differentiating through the base-level learner's optimisation process using automatic differentiation tools. Gradient signals are propagated not only through time, but also across the recursive learning hierarchy, enabling a principled and data-driven approach to discovering update rules.


A concrete realisation of the policy loss function learned by the meta-learner is given
\[
\mathcal{L}_{\text{policy}}^\Phi(\theta, \tau) = \mathbb{E}_{(s, a, r, s') \in \tau} \left[ -\hat{A}^\Phi(s, a) \log \pi_\theta(a|s) + \lambda^\Phi(s) \cdot \mathcal{H}[\pi_\theta(\cdot|s)] \right],
\]
where the advantage estimate $\hat{A}^\Phi$ and entropy coefficient $\lambda^\Phi$ are outputs of neural networks parameterised by $\Phi$. This formulation generalises commonly used actor-critic methods and allows the meta-learner to shape the reward and exploration structures directly.

The recursive structure of the framework necessitates propagating gradients through entire optimisation procedures. This is achieved using automatic differentiation frameworks which support both unrolled gradient computation and implicit differentiation. In unrolled differentiation, the full sequence of parameter updates is stored and backpropagated through, while in the implicit case, gradients are computed by differentiating fixed-point optimality conditions, thus saving memory and improving scalability.

This instantiation of our framework provides a mechanism to learn policy update rules from data. By allowing both the form and content of the objective function to be learned through nested optimisation, and by enabling gradient flow through all layers of abstraction, this approach opens a new direction in reinforcement learning. In particular, it offers the potential to discover novel and problem-specific update rules that generalise beyond standard approaches, all while preserving differentiability and enabling integration with existing learning pipelines.

\end{tcolorbox}

\section*{Degenerate Cases} An important property of the proposed framework is that it subsumes many well-established machine learning techniques as degenerate cases. This both grounds the framework in existing practice and highlights its generality.

In the simplest case, when the recursion depth is set to $K = 1$ and no virtual tasks are used, our framework reduces to standard meta-learning algorithms such as MAML~\cite{finn2017model} or Reptile~\cite{nichol2018reptile}. Here, the meta-learner optimises an initialisation or learning rule for a set of tasks sampled from a distribution, with no further abstraction. When $K = 1$ is retained but virtual tasks are introduced through synthetic data generation, our framework recovers known techniques in data augmentation and consistency regularisation. For instance, approaches such as SimCLR~\cite{chen2020simple} and FixMatch~\cite{sohn2020fixmatch} rely on augmenting the training data with synthetic variants, effectively introducing a rudimentary form of virtual task generation. Increasing the depth to $K = 2$ and introducing learned soft constraints connects our framework to meta-regularisation methods. In this setting, the meta-learner learns to guide the training of lower-level learners by shaping their loss landscapes, akin to techniques such as Learning to Learn Reinforcement Learn~\cite{wang2016learning} and meta-learned regularisation~\cite{balaji2018metareg}. 

Finally, recursive application of our framework without explicit virtual task generation corresponds naturally to curriculum learning~\cite{hacohen2019power,graves2017automated}. Here, each meta-level abstracts over the one below, progressively learning higher-level representations and training curricula, echoing the philosophy of self-paced learning.
These connections illustrate that the proposed framework provides a unifying structure in which existing approaches can be understood as instances of a broader, compositional approach to learning. This perspective is particularly valuable when designing novel learning pipelines that combine elements of meta-learning, curriculum learning, and adversarial training.

\section*{Conclusion}

We propose a unified framework for higher-order meta-learning using category theory and PINN-inspired soft constraints. Virtual tasks play the role of collocation points, enforcing meta-properties at each level. Meta-learners are promoted recursively to higher-order learners, enabling the design of deep, structured meta-learning pipelines with explicit control over both learning dynamics and meta-properties. The benefits of this approach are numerous. Firstly, it enables the design of \emph{deeply structured meta-learning pipelines}, where each layer can explicitly learn to improve the learning dynamics of the layer below. Secondly, the use of category theory provides a unifying language for reasoning about these complex compositions of learners, offering both conceptual clarity and practical tools for modular implementation. Thirdly, the incorporation of soft constraints allows for the enforcement of desirable properties, such as invariance, robustness, and interpretability, in a flexible and data-efficient manner. Importantly, this framework also facilitates transfer learning and continual learning scenarios, as the constraints and meta-learning objectives can naturally encode knowledge about task families and learning dynamics.Overall, this recursive meta-learning architecture extends the flexibility and power of modern neural networks to the meta-learning domain, enabling the systematic and principled design of learning processes that themselves learn and adapt. The analogy to PINNs provides a practical template for implementation.
\\\textbf{Meta-Cognition, Self-Awareness, and Black-Box Learning.} The recursive meta-learning framework described in this work exhibits computational mechanisms that bear strong resemblance to forms of \textit{meta-cognition}~\cite{rhodes2019metacognition,schwarz2015metacognition} and \textit{self-awareness} \cite{wicklund1975objective,gallup1998self,gallup1982self}, as understood in cognitive psychology. At each level of the hierarchy, a meta-learner explicitly monitors and evaluates the performance of learners at subordinate levels across varying task instances. When the meta-learner observes poor generalisation or high loss on tasks that are structurally near others where the learner performs well, this signals a failure of inductive abstraction---analogous to a cognitive agent recognising a limitation in its current conceptual understanding. In response, the framework generates targeted virtual tasks that probe these limitations and refine soft constraints to support more general learning. This recursive introspection and adaptation mimics key features of what psychologists term \textit{meta-cognitive awareness}, the ability to recognise the boundaries of one's own competence and to intentionally engage in behaviour that improves performance over time~\cite{schraw1995metacognitive,schraw1994assessing}.

This process also connects to the notion of \textit{self-efficacy};the belief in one's capacity to learn or solve a problem in the sense that the system develops internal models of capability~\cite{gallup1998self}. By exploring the difficulty landscape and synthesising increasingly general tasks, the meta-learner effectively shapes its own curriculum, progressively building more abstract representations of problem classes and developing strategies to address tasks it previously failed. In doing so, it not only reacts to failure but also systematically refines the space of what it believes can be learned, and how. Importantly, this recursive structure offers a potential bridge between high-level conceptual learning and the low-level computation often viewed as a \textit{black box} in deep learning models. Because each layer explicitly conditions the next through soft constraints and generates diagnostic virtual tasks, the system creates a traceable path of influence between failure signals, constraint adaptation, and eventual improved performance. This chain of reasoning can be used to interpret why particular inductive biases emerge and how knowledge is generalised. Moreover, by analysing the virtual task generation mechanism and the evolving constraint manifolds at each level, one gains insight into the internal epistemology of the learner; its representation of task-space, its beliefs about generalisation, and its dynamic estimation of its own competence. These features suggest that such architectures may serve as interpretable models of artificial systems that are aware of their own learning processes, and capable of self-improvement grounded in structural awareness.

\bibliographystyle{alpha}
\bibliography{sample}




\clearpage
\appendix
\section{Mathematical Structures for Virtual Manifold} \label{sec:maths_structure}
\textbf{Rings and Modules.} A \emph{ring} \( R \) is an algebraic structure equipped with two operations; addition and multiplication, that generalise integer arithmetic. If the constraint manifold involves composition rules or additive constraints (e.g., cost accumulation, abstract grammars), representing elements as members of a ring or a module over a ring can enable tractable modelling of these operations. 

\noindent\textbf{Lie groups.} Lie groups are smooth manifolds equipped with a group structure, where the group operations (composition and inversion) are differentiable. They arise naturally in contexts where constraints exhibit continuous symmetries, such as rotations, translations, and scalings. Formally, a Lie group \( G \) is both a group and a smooth manifold such that the maps \( G \times G \to G \), \( (g, h) \mapsto gh^{-1} \), are smooth. In the context of learned constraint manifolds, Lie groups can be used to model symmetries that is, they can ensure that constraints are preserved under group actions \( g \cdot \tau \) for \( g \in G \), where \( \tau \) denotes a task, embed equivariance, that is design constraint networks or generators whose outputs transform equivariantly under group actions.

\subsection*{Smooth Contextual Exploration} \label{sec:smooth_context}
To enable stable gradient flow and efficient optimisation during exploration, we introduce a smooth approximation of the contextual exploration score. Algorithm \ref{alg:recursive-meta_adv_gen} relied on non-differentiable \texttt{max} operations, which are replaced here by the \textit{softplus} function to preserve the semantics of threshold-based scoring while ensuring differentiability. The resulting criterion rewards tasks that exhibit either (i) significant performance deviation from nearby tasks or (ii) unexpectedly similar performance to distant tasks. These terms encourage exploration of sharp transitions and surprising generalisation patterns respectively.

We define the smooth contextual exploration score as
\begin{align*}
\mathcal{S}_{\text{explore}}^{\text{smooth}}(\tau) =\; & \lambda \cdot \text{softplus}\left( 
\frac{\Delta \mathcal{L}(\tau_\text{ref}, \tau)}{d(\tau, \tau_{\text{ref}}) + \delta} - \epsilon_1 - \alpha_1
\right) + \alpha_1 \\
& + (1 - \lambda) \cdot \text{softplus}\left( 
\frac{d(\tau, \tau_{\text{ref}})}{1 + \left| \Delta \mathcal{L}(\tau_\text{ref}, \tau) \right|} - \epsilon_2 - \alpha_2
\right) + \alpha_2,
\end{align*}
where,  $\alpha_1, \alpha_2 \ll 0$ set the baseline values for the \texttt{softplus}-based thresholding. The term $\delta > 0$ avoids numerical instability near zero distance. This formulation provides a smooth, tunable mechanism for encouraging exploration over the task space that remains sensitive to the learner's loss landscape. While effective, the use of extremal reference tasks can be brittle due to outliers or noise. We therefore introduce a smoothed variant that soft-aggregates past task information via kernel-weighted averages. Let $K(\tau, \tau')$ be a positive semi-definite kernel (e.g., Gaussian) that quantifies similarity between tasks $\tau$ and $\tau'$. We define
\begin{align*}
\overline{\mathcal{L}}_{\text{ref}}(\tau) = \frac{\sum_{\tau' \in \mathcal{H}} K_{\text{ref}}(\tau, \tau') \, \mathcal{L}(\theta, \tau')}{\sum_{\tau' \in \mathcal{H}} K_{\text{ref}}(\tau, \tau')}, \quad
\overline{d}_{\text{ref}}(\tau) = \frac{\sum_{\tau' \in \mathcal{H}} K_{\text{ref}}(\tau, \tau') \, d(\tau, \tau')}{\sum_{\tau' \in \mathcal{H}} K_{\text{ref}}(\tau, \tau')} 
\end{align*}

The full kernel-smoothed contextual exploration score is
\begin{align*}
\mathcal{S}_{\text{explore}}^{\text{kernel}}(\tau) =\; & \lambda \cdot \text{softplus} \left(
\frac{\mathcal{L}(\theta, \tau) - \overline{\mathcal{L}}_{\text{ref}}(\tau)}{\overline{d}_{\text{ref}}(\tau) + \delta} - \epsilon_1 - \alpha_1
\right) + \alpha_1 \\
& + (1 - \lambda) \cdot \text{softplus} \left(
\overline{\Delta d}_{\text{ref}}(\tau) - \epsilon_2 - \alpha_2
\right) + \alpha_2,
\end{align*}
where
$K_{\text{ref}}(\tau, \tau') = \exp\left(-\gamma_{\text{ref}} d(\tau, \tau')^2\right)$ concentrates on nearby tasks
.
This kernel-smoothed approach aggregates past experiences in a locality-sensitive manner, reducing variance in exploration signals while preserving contextual diversity. It also provides gradient flow through the full task history, making it particularly suitable for memory-based exploration in generative models. It is particularly advantageous in high-dimensional task spaces where sparsity of visits or noise may make extremal points unreliable. By incorporating both sharp local transitions and global generalisation, this criterion ensures that generated tasks remain informative throughout training. We embed the kernel-smoothed contextual exploration score into the generator's training objective. Let $G_\phi(z)$ denote the generator network parameterised by $\phi$, which maps latent codes $z \sim \mathcal{Z}$ to synthetic tasks $\tilde{\tau} = G_\phi(z)$. The generator is optimised to maximise the expected exploration score over generated tasks $
\mathcal{L}_{\text{gen}}(\phi) = - \mathbb{E}_{z \sim \mathcal{Z}} \left[
\mathcal{S}_{\text{explore}}^{\text{kernel}}(G_\phi(z))
+ \beta \mathcal{L}_{\text{manifold}}(G_\phi(z))
- \gamma \log D_\psi(G_\phi(z))\right]$,
where the parameters $\phi$ are updated via stochastic gradient ascent to maximise this loss. 
\vspace{1cm}





\footnotesize
\begin{algorithm}[H]
\caption{\small Sample-Efficient Recursive Meta-Learning}
\label{alg:sample_efficient_meta_learning}
\KwIn{Meta-levels $K$, task distributions $\{\mathcal{D}_k\}$, initial parameters $\{\Phi_k\}$, memory buffers $\{\mathcal{M}_k\}$}
\KwOut{Optimised base-level objective or learner}

\For{$k \gets K$ \KwTo $1$}{
    \ForEach{batch of meta-tasks $\mathcal{T}_k \sim \mathcal{D}_k$}{
        Schedule task difficulty using curriculum controller at level $k$\;
        Retrieve relevant prior experience from memory buffer $\mathcal{M}_k$\;
        Initialise or sample meta-learner at level $k-1$ with parameters $\Phi_{k-1}$\;

        \ForEach{task $\mathcal{T} \in \mathcal{T}_k$}{
            \ForEach{task instance $\tau \in \mathcal{T}$}{
                Initialise learner $f_\theta$ (e.g., policy or classifier)\;
                Compute surrogate loss $\hat{\mathcal{L}}^{\Phi_{k-1}}(\theta, \tau)$ using differentiable objective conditioned on $\Phi_{k-1}$\;
                \tcp{Backpropagate through unrolled learning steps}
                Update $\theta$ with gradient descent: $\theta \gets \theta - \alpha \nabla_\theta \hat{\mathcal{L}}^{\Phi_{k-1}}(\theta, \tau)$\;
                Store training trace and performance metrics in buffer $\mathcal{M}_k$\;
                Evaluate performance $R(\theta, \tau)$ and compute task-specific weight $w(\tau)$\;
            }

            \If{virtual task generation is required}{
                Sample synthetic task $\tilde{\tau} \sim G_{\phi_k}(z)$\;
                Optionally train generator $G_{\phi_k}$ adversarially with discriminator $D_{\psi_k}$\;
            }

            Compute virtual surrogate loss:
            \[
            \mathcal{L}_{\text{virtual}}^{\Phi_k} = \sum_{\tilde{\tau}} w(\tilde{\tau}) \cdot \hat{\mathcal{L}}^{\Phi_{k-1}}(\theta, \tilde{\tau})
            \]
        }

        Compute regularised meta-objective:
        \[
        \mathcal{L}_{\text{meta}}^{\Phi_k} = \sum_{\mathcal{T}} R(\theta_{\mathcal{T}}) + \lambda \mathcal{L}_{\text{virtual}}^{\Phi_k} + \beta \mathcal{R}(\Phi_k)
        \]

        Update meta-learner parameters:
        \[
        \Phi_k \gets \Phi_k - \eta \nabla_{\Phi_k} \mathcal{L}_{\text{meta}}^{\Phi_k}
        \]
    }
}
\Return{Optimised learner or objective at base level ($k=0$)}
\end{algorithm}
\normalsize
\section{Game Theory: Background}
A \textit{normal form game} \cite{osborne1994course} with \( N \) players is defined as $
G = (A_1, \dots, A_N; u_1, \dots, u_N)$,
where each player \( i \in \{1, \dots, N\} \) has a finite action set \( A_i \), and a payoff function $
u_i : A_1 \times \dots \times A_N \rightarrow \mathbb{R}$. A \textit{Nash equilibrium} is a strategy profile \( (\sigma_1^*, \dots, \sigma_N^*) \in \Delta(A_1) \times \dots \times \Delta(A_N) \) such that for each player \( i \), $
\sigma_i^* \in \arg\max_{\sigma_i \in \Delta(A_i)} \mathbb{E}_{a \sim \sigma_{-i}^* \times \sigma_i}[u_i(a)],$
where \( \sigma_{-i}^* \) denotes the strategy profile of all players except player \( i \), and \( a = (a_1, \dots, a_N) \sim \sigma_1 \times \dots \times \sigma_N \) denotes a joint action sampled from the product distribution over the players’ mixed strategies. 

\end{document}